\documentclass[journal]{IEEEtran}
\usepackage{subfig}

\usepackage{times}
\usepackage{epsfig}
\usepackage{graphicx}
\usepackage{amsmath}
\usepackage{amssymb}

\usepackage{booktabs}       
\usepackage{amsfonts}       
\usepackage{nicefrac}       
\usepackage{microtype}      
\usepackage{bm}
\usepackage{myothers}
\usepackage{algorithm}
\usepackage{algorithmicx}

\usepackage{mathdots}
\usepackage{mdwmath}
\usepackage{mdwtab}
\usepackage{paralist}
\usepackage{color}
\usepackage{bbding}

\usepackage[pagebackref=true,breaklinks=true,letterpaper=true,colorlinks,bookmarks=false]{hyperref}

\ifCLASSINFOpdf
\else
\fi
\begin{document}
%
\title{Fast Kernelized Correlation Filter \\
without Boundary Effect}
%
%
%

\author{Ming~Tang\textsuperscript{*},~\IEEEmembership{Member,~IEEE,}
        Linyu~Zheng\textsuperscript{*}, Bin Yu,
        and~Jinqiao~Wang,~\IEEEmembership{Member,~IEEE}
\thanks{The authors are with the National Lab of Pattern Recognition, Institute of Automation, Chinese Academy of Sciences (Beijing 100190), and School of Artificial Intelligence, University of Chinese Academy of Sciences (Beijing 100049), China.}
\thanks{* Indicates joint first author.}
\thanks{The corresponding author is Ming Tang (tangm@nlpr.ia.ac.cn)}
\thanks{This paper has supplementary downloadable material available at
http://ieeexplore.ieee.org., provided by the authors. The material includes an illustration to explain how Algorithm 2, CCIM, works exactly and a mathematical proof of the correctness of CCIM. Contact the corresponding author for further questions about this work.}
\thanks{Manuscript received Match xx, 2020. 
}
}
\maketitle

\begin{abstract}
In recent years, correlation filter based trackers (CF trackers) have attracted much attention from the vision community because of their top performance in both localization accuracy and efficiency. The society of visual tracking, however, still needs to deal with the following difficulty on CF trackers: avoiding or eliminating the boundary effect completely, in the meantime, exploiting non-linear kernels and running efficiently. In this paper, we propose a fast kernelized correlation filter without boundary effect (nBEKCF) to solve this problem. To avoid the boundary effect thoroughly, a set of \emph{real} and \emph{dense} patches is sampled through the traditional sliding window and used as the training samples to train nBEKCF to fit a Gaussian response map. Non-linear kernels can be applied naturally in nBEKCF due to its different theoretical foundation from the existing CF trackers'. To achieve the fast training and detection, a set of cyclic bases is introduced to construct the filter. Two algorithms, ACSII and CCIM, are developed to significantly accelerate the calculation of kernel correlation matrices. ACSII and CCIM fully exploit the density of training samples and cyclic structure of bases, and totally run in space domain. The efficiency of CCIM exceeds that of the FFT counterpart remarkably in our task. Extensive experiments on six public datasets, OTB-2013, OTB-2015, NfS, VOT2018, GOT10k, and TrackingNet, show that compared to the CF trackers designed to relax the boundary effect, BACF and SRDCF, our nBEKCF achieves higher localization accuracy without tricks, in the meanwhile, runs at higher FPS.
\end{abstract}


%
\IEEEpeerreviewmaketitle

\section{Introduction}

\begin{figure*}[t]
  \begin{center}
  \subfloat[]{
  \label{fig:illus-nbekcf2}
  \includegraphics[width=0.45\textwidth]{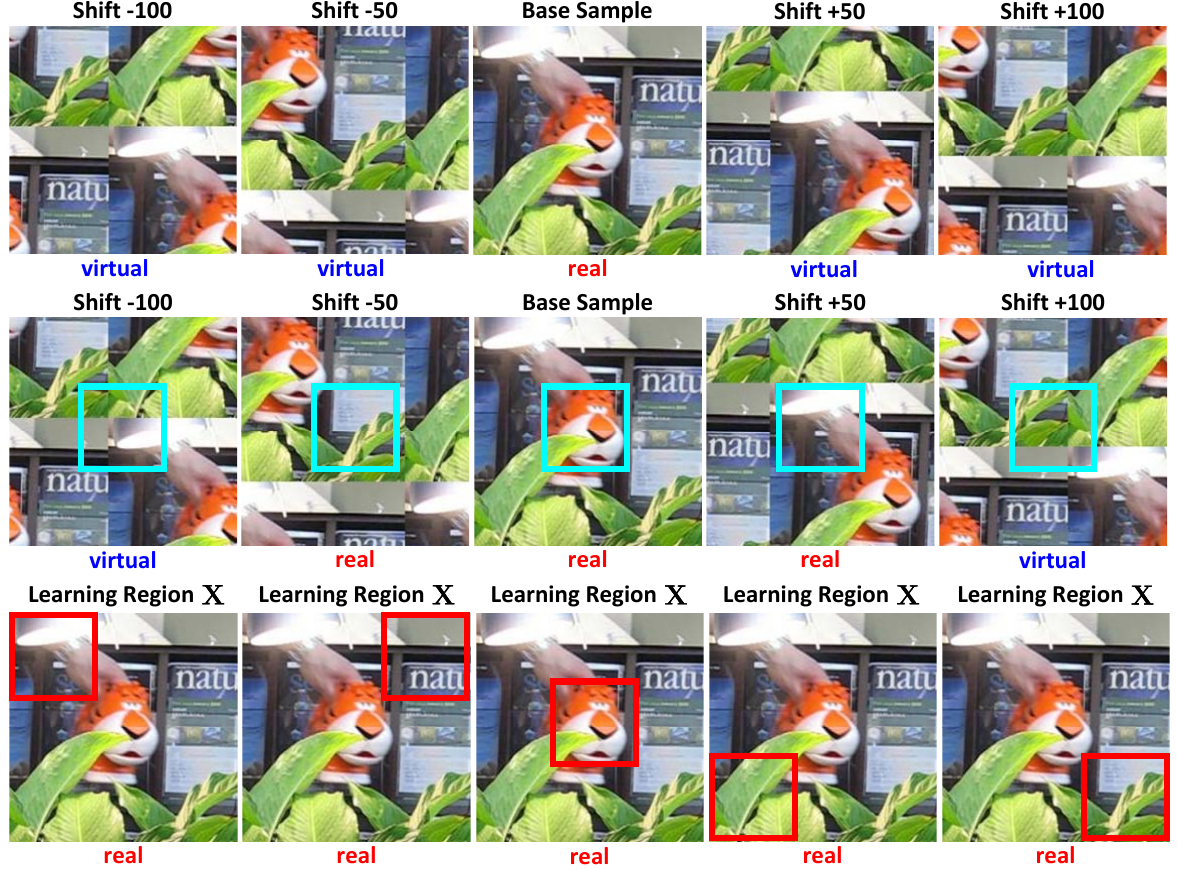}}
  \hskip 8mm
  \subfloat[]{
  \label{fig:illus-nbekcf1}
  \includegraphics[width=0.45\textwidth]{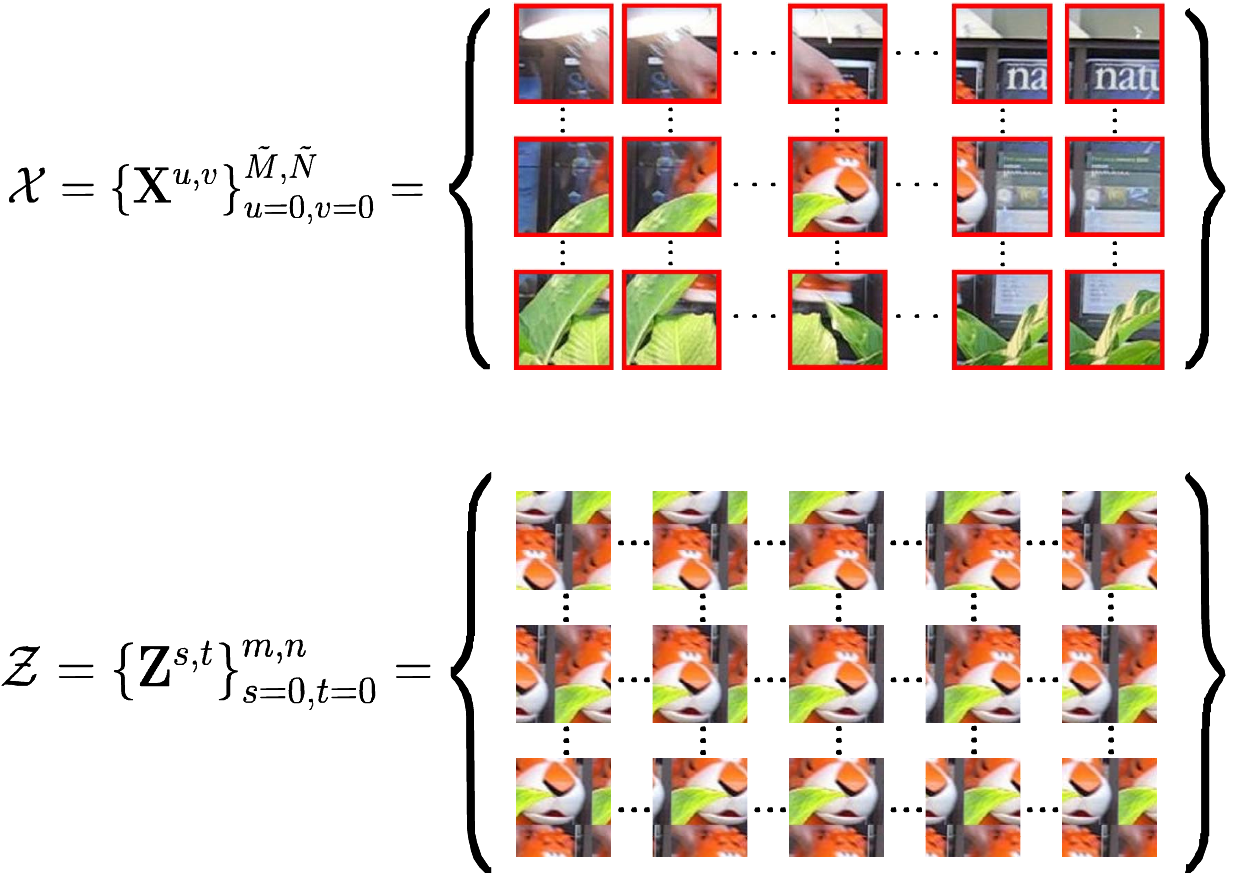}}
  \end{center}
  \caption{(a) Comparison of sampling methods in KCF~\cite{henriques15} (the first row), BACF~\cite{galoo17a} (the second row), fdKCF*~\cite{fdkcf-zhengly19} and our nBEKCF (the last row). Training samples of KCF come from all possible cyclic shifts of a base sample (\ie, the central patch), and they are all virtual except for the base one. So do those of SRDCF~\cite{dane15a} and ECO~\cite{dane17b}. BACF obtains its training samples of target size (cyan boxes) by clipping the middle parts of all training samples of KCF, therefore, some of them are virtual. Different from KCF and BACF, in fdKCF* and our nBEKCF, the training samples of target size (red boxes) are densely sampled from the learning region $\mathbf{X}$ with the traditional sliding window, and they are all real. We call such sampling method as \emph{real and dense sampling}. (b) Illustrations of a training set $\mathcal{X}$ and a set $\mathcal{Z}$ of cyclic bases in nBEKCF. $\mathbf{X}^{u,v}$'s are sampled from $\mathbf{X}$ by using the real and dense sampling method. The elements, $\mathbf{Z}^{s,t}$'s, of $\mathcal{Z}$ are pre-defined and totally \emph{cyclic}. $\mathcal{Z}$ is constructed by \emph{all possible cyclic shifts} of the target patch (\ie, the central red box on the last row of (a)) in $\mathbf{X}$, although theoretically it is not necessary to use target patches to generate $\mathbf{Z}^{s,t}$'s. Note that both the density of $\mathcal{X}$ and the cyclicity of $\mathcal{Z}$ are crucial to the high efficiency of our nBEKCF.}
\label{fig:illus-nbekcf}
\end{figure*}

Visual tracking is one of the fundamental problems in computer vision with many applications.
Despite significant progress in recent years~\cite{vot2015,vot2016,vot2017,vot2018,li2018deep,vot2019}, visual tracking is still a challenge~\cite{wu15} due to some severe interferences (e.g. large appearance changes, occlusions, background clutters and fast motion), very limited training samples, and the requirement of low computational cost. In tracking task, it is crucial to construct a robust appearance model from very limited samples to distinguish a target from distractive background, while maintaining high efficiencies.

Since 2010, correlation filter based trackers (CF trackers) have been achieving a great success~\cite{bolme10,henriques2012,dane15a,tangm15,dane16b,galoo17a,dane17b,bhat18,tangm18,
sunc18,fdkcf-zhengly19}. Almost all CF trackers learn their filters with cyclic samples to regress Gaussian response maps, and their training and detection are accelerated with the convolution theorem and fast Fourier transform (FFT). Bolme~\etal~\cite{bolme10} proposed the Minimum Output Sum of Squared Error (MOSSE) for very high speed tracking on gray-scale sequences. They used base image patches and all their cyclical shifts to train the appearance model directly in Fourier domain. Henriques~\etal~\cite{henriques15} reformulated MOSSE as a ridge regression problem in space domain, being able to apply multi-channel features and non-linear kernels naturally to improve the localization accuracy. In their CF tracker, kernelized correlation filter (KCF), the regression function and its optimization problem are expressed as
\begin{equation}
\label{eq:kcf-regression}
f(\mathbf{X})=\sum_{s=0}^{m-1}\sum_{t=0}^{n-1}\alpha_{s,t}\kappa(\mathbf{X},\mathbf{X}_{s,t})
\end{equation}
and
\begin{equation}
\label{eq:kcf-objfunction}
\min_{\bm{\alpha}}\|\mathbf{K}\bm\alpha-\mathbf{y}\|_2^2+\lambda\bm\alpha^{\top}\mathbf{K}\bm\alpha,
\end{equation}
respectively, where $\mathbf{X}$ and $\mathbf{X}_{s,t}$'s are cyclic sample, and $\mathbf{K}$ is the kernel matrix with $\kappa(\mathbf{X}_{u,v},\mathbf{X}_{s,t})$ as its elements, $\{\mathbf{X}_{u,v}\}\equiv\{\mathbf{X}_{s,t}\}\equiv\mathcal{X}_{\text{KCF}}$, $\mathcal{X}_{\text{KCF}}$ is the set of cyclic samples. Compared to MOSSE, KCF achieves a much higher accuracy on OTB-2013~\cite{wu13} when exploiting HOG feature~\cite{dalal2005} and Gaussian kernel, meanwhile, it is still able to run at a high speed.
However, the use of FFT produces the cyclicity of samples which leads to the problem of \emph{boundary effect}~\cite{galoo15} in MOSSE, KCF, and many CF trackers~\cite{tangm15,luke17,mueller17,valmadre17,zhang2018learning,tangm18}. As shown in Fig.~\ref{fig:illus-nbekcf2}, the boundary effect means that almost all training and detection samples are unreal and synthesized by cyclically shifting base samples, and these unreal samples are still supposed to represent those real ones at different translational shifts in training and localization.\footnote{Readers may refer to the $4^{\text{th}}$ paragraph of Sec.1 in~\cite{galoo15} for other details.} As for the training of KCF, all training samples, except for a real base sample, \ie, $\mathbf{X}_{0,0}$, are unreal in Problem (\ref{eq:kcf-objfunction}). In practice, the boundary effect dramatically reduces the discriminative power of appearance models and greatly degrades the localization accuracy which MOSSE and KCF could have achieved.

In order to relax the boundary effect to improve localization accuracy of the KCF with linear kernel, Galoogahi~\etal~\cite{galoo17a} and Danelljan~\etal~\cite{dane15a} proposed the background-aware correlation filter (BACF) and spatially regularized discriminative correlation filter (SRDCF), respectively. In BACF, a rectangular mask is introduced into the error item of Problem (\ref{eq:kcf-objfunction}) to cover the cyclic samples, and then the alternating direction method of multipliers (ADMM) is employed to solve the optimization problem with equality constraints. Although the boundary effect can be reduced greatly by introducing the mask, it cannot be eliminated completely in BACF, as pointed out in~\cite{galoo15} and~\cite{on-relations-wangjq19}. In SRDCF, a smooth spatial regularization factor is introduced into the regularizer of Problem (\ref{eq:kcf-objfunction}) to penalize the filter coefficients depending on their spatial locations. The regularization factor acts really similarly to the mask of BACF in practice~\cite{on-relations-wangjq19}, being not able to eliminate the boundary effect thoroughly. On the other hand, due to the ways they relax the boundary effect, SRDCF and BACF cannot run in high efficiencies when employing the powerful high-dimensional features. Although it has been shown that trackers can improve their accuracies by large margins with non-linear kernels~\cite{zuo-pami19}, SRDCF and BACF are also unable to exploit the non-linear kernels to improve their localization accuracy because the window shape is unknown in the non-linear kernel space, despite the fact that it is known in an image or filter with linear kernel.

The latest development of SRDCF is UPDT~\cite{bhat18}. UPDT applies the same way as SRDCF's to relax the boundary effect, and achieves the state-of-the-art accuracy through the following series of improvements, \ie, (1) the interpolation of feature resolution which was designed to improve SRDCF and resulted in C-COT~\cite{dane16b}; (2) the reduction of feature dimensionality, linear weighting of features, clustering samples, and sparse update which were developed to improve C-COT and resulted in ECO~\cite{dane17b}; (3) the employment of deeper networks, ResNet50, the augmentation of training samples, and the adaptive fusion of models with shallow and deep features which were designed to improve ECO. Therefore, UPDT inherits various defects of SRDCF on the boundary effect.

According to the above statements, it is seen that there still exists the following problem for CF trackers: avoiding or eliminating the boundary effect completely, and in the meantime, exploiting non-linear kernels and running in a high efficiency.

In fact, the efficiencies of MOSSE and KCF totally rely on the boundary effect. Therefore, the efforts of BACF and SRDCF to reduce the boundary effect on the theoretical foundation of KCF is bound to weaken their efficiencies seriously. On the other hand, as a side-effect, BACF and SRDCF's efforts to reduce the boundary effect inhibit them from applying non-linear kernels to improve their accuracies. According to this analysis, we believe that a totally novel theoretical foundation other than that of KCF is necessary to develop a novel type of correlation filter to address the above problem.

Now that both the efficiency and the boundary effect of KCF is from the cyclicity of $\mathcal{X}_{\text{KCF}}$, if $\{\mathbf{X}_{u,v}\}$ and $\{\mathbf{X}_{s,t}\}$ are treated as two different sets, \ie, treat $\{\mathbf{X}_{u,v}\}$ as a training set and $\{\mathbf{X}_{s,t}\}$ as a base set, the cyclicity of $\{\mathbf{X}_{u,v}\}$ is cancelled, \ie, all $\mathbf{X}_{u,v}$'s are real, and the cyclicity of $\{\mathbf{X}_{x,t}\}$ is kept, then the boundary effect of KCF disappears and the efficient calculation is still able to achieved. According to this idea, in this paper, we propose a novel type of CF tracker, a fast kernelized correlation filter without boundary effect (nBEKCF), which is totally different from KCF in theory, to solve the above problem. 
As shown in Fig.\ref{fig:illus-nbekcf2}, unlike most existing CF trackers which exploit both real and synthetic patches generated from a base image patch as training samples to train their filters with FFT, our nBEKCF draws a set of training samples by using the real and dense sampling method to fit a Gaussian response map, avoiding the boundary effect thoroughly. To train the filter and locate the target object efficiently, a set of cyclic bases is introduced and exploited without FFT. Specifically, a set of basis functions, $\{\kappa(\cdot,\mathbf{Z}^{s,t})\}$, is generated with a set of pre-defined cyclic bases $\mathcal{Z}=\{\mathbf{Z}^{s,t}\}$, and our filter is formulated with this basis function set, \ie, $f(\cdot;\mathbf{Z}^{s,t})=\sum_{s,t}\alpha_{s,t} \kappa(\cdot,\mathbf{Z}^{s,t})$. Training set $\mathcal{X}$ consists of real $\mathbf{X}^{u,v}$'s which are \emph{densely sampled} from learning region $\mathbf{X}$ in a pixel-wise way. Fig.\ref{fig:illus-nbekcf1} illustrates training sets $\mathcal{X}$ and $\mathcal{Z}$. Then, filter $f(\cdot;\mathbf{Z}^{s,t})$ regresses training set $\mathcal{X}=\{\mathbf{X}^{u,v}\}$ to a Gaussian response map. Note that the non-linear kernels can be applied naturally in $f(\cdot;\mathbf{Z}^{s,t})$. In order to treat multiple frames and update the filter efficiently, the modeling scheme over multiple frames~\cite{dane14a,tangm18} is adapted to nBEKCF. 
It is worth noticing that $\mathcal{X}\neq\mathcal{Z}$ and only $\mathcal{Z}$ is cyclic in nBEKCF, but $\mathcal{X}\equiv\mathcal{Z}\equiv\mathcal{X}_{\text{KCF}}$ and $\mathcal{X}_{\text{KCF}}$ is cyclic in KCF.

It is found that the key to improve the efficiency of training and detection is the quick calculation of kernel correlation matrices in nBEKCF. Therefore, we develop two \emph{non-FFT} based algorithms, autocorrelation with squared integral image (ACSII) and cyclic correlation with integral matrix (CCIM), to significantly accelerate the calculation by fully exploiting the density of $\mathcal{X}$ and cyclic structure of $\mathcal{Z}$. In our approach, a kernel correlation matrix is constructed with $\kappa(,)$, $\mathcal{Z}$ and $\mathcal{X}$. By exploiting a great deal of overlap among densely sampled $\mathbf{X}^{u,v}$'s, ACSII calculates the autocorrelation efficiently. By exploiting both the density of $\mathbf{X}^{u,v}$'s and the cyclicity of $\mathbf{Z}^{s,t}$'s, CCIM is remarkably more efficient than that of FFT in calculating the correlation in our task.

Consequently, the regression problem is solved efficiently in space domain, rather than by means of frequency domain, in nBEKCF.
It is $\mathcal{X}\neq\mathcal{Z}$ and all elements of $\mathcal{X}$ being real that make nBEKCF free from the boundary effect, and it is $\mathcal{X}\neq\mathcal{Z}$ and the cyclicity of $\mathcal{X}$ that make nBEKCF locate the target object really efficiently.




Our nBEKCF is tested on six public datasets, OTB-2013, OTB-2015, NfS, VOT2018, GOT10k, and TrackingNet. The experimental results show that nBEKCF achieves state-of-the-art accuracy, and compared to the trackers designed to relax the boundary effect, BACF and SRDCF, our nBEKCF with hand-crafted features, HOG and CN, obtains higher localization accuracy without tricks and is able to run at higher FPS (50 on average).

In summary, our main contributions are as follows.
\begin{itemize}
\item[1)] We propose a totally novel design theory of correlation filter, and develop a novel type of CF tracker entitled nBEKCF based on this theory. nBEKCF avoids the boundary effect thoroughly as well as is able to exploit non-linear kernels naturally and can run in a high efficiency.

\item[2)] We develop an efficient algorithm, CCIM, to fully exploit both the density of training samples and the cyclicity of bases to construct kernel correlation matrices of nBEKCF. CCIM runs in space domain, and exceeds FFT significantly in efficiency in our task.

\item[3)] We design an algorithm, ACSII which runs in space domain, to fully exploit the density of training samples to fast calculate the autocorrelation in nBEKCF.
\end{itemize}



\section{Kernelized Correlation Filter without Boundary Effect (nBEKCF)}
\label{sec:hskcfwobe}
In this section, we will first introduce our novel nBEKCF with a single frame, then extend it to the historical frames, and finally present how to locate the target object with nBEKCF in the current frame.


\subsection{nBEKCF with Single Frame}
\label{sec:hskcfwobes}

Let $\mathbf{Z}\in\mathbb{R}^{m\times n\times D}$ be the $D$-channel feature map of the base patch. In our current implementation, the base patch is the target object patch. The set of cyclic bases, $\mathcal{Z}=\left \{ \mathbf{Z}^{s,t} \right \}_{s=0, t=0}^{m-1, n-1}$, is generated by
\begin{equation}
\label{eq:cycliccfs}
\mathbf{Z}^{s,t}\left ( d \right )=\mathbf{P}_m^{s}\mathbf{Z}\left ( d \right )\mathbf{Q}_n^{t},
\end{equation}
where $d=0,\ldots,D-1$, $\mathbf{Z}\left ( d \right )$ is the $d$-th channel of $\mathbf{Z}$, $\mathbf{P}_m$ and $\mathbf{Q}_n$ are the $m\times m$ and $n\times n$ permutation matrices~\cite{davidp94}, respectively,
\begin{equation*}
\mathbf{P}_{m}=\left[\begin{array}{cc}
  \mathbf{0}_{m-1}^{\top} & 1\\
  \mathbf{I}_{m-1} & \mathbf{0}_{m-1}\\
  \end{array}
  \right],\;\;
\mathbf{Q}_{n}=\left[\begin{array}{cc}
  \mathbf{0}_{n-1} & \mathbf{I}_{n-1}\\
  1 & \mathbf{0}_{n-1}^{\top}\\
  \end{array}
  \right],
\end{equation*}
where $\mathbf{0}_{l-1}$ is the $(l-1)\times 1$ zero vector, $\mathbf{I}_{l-1}$ is the $(l-1)\times (l-1)$ identity matrix, $l\in \left \{ m,n \right \}$, and $\mathbf{P}_{m}^{\rho}$ and $\mathbf{Q}_{n}^{\rho}$ are the $\rho$-th power of $\mathbf{P}_{m}$ and $\mathbf{Q}_{n}$, respectively. Note that $\mathbf{Z}^{0,0}(d)=\mathbf{Z}(d)$. Intuitively, $\mathbf{P}_{m}^{\rho}\mathbf{Z}(d)$ cyclically shifts $\mathbf{Z}(d)$'s rows down by $\rho$ rows and $\mathbf{Z}(d)\mathbf{Q}_{n}^{\rho}$ cyclically shifts $\mathbf{Z}(d)$'s columns right by $\rho$ columns, when $\rho\geqslant0$.\footnote{If $\rho<0$, the direction of shift is opposite to that of $\rho>0$.} Therefore, $\mathbf{Z}^{0,0}=\mathbf{Z}$, and $\mathcal{Z}$ consists of all possible cyclic shifts of $\mathbf{Z}$ on 2D spatial domain.

Furthermore, let $\mathbf{X}\in\mathbb{R}^{M\times N\times D}$ be the $D$-channel feature map of learning region. We generate the set of real (\ie, non-cyclic) training samples of target size, $\mathcal{X}=\left \{\mathbf{X}^{u,v}\in\mathbb{R}^{m\times n\times D}\right \}_{u=0,v=0}^{M-m,N-n}$, through real and dense sampling from $\mathbf{X}$, as shown in Fig.\ref{fig:illus-nbekcf}.


Then, let $f:\mathbb{R}^{m\times n\times D}\rightarrow\mathbb{R}$, kernel $\kappa:\mathbb{R}^{m\times n\times D}\times\mathbb{R}^{m\times n\times D}\rightarrow\mathbb{R}$, and consider $\{\kappa(\cdot,\mathbf{Z}^{s,t})\}_{s=0,t=0}^{m-1,n-1}$ as a set of basis functions. We define the kernelized correlation filter without boundary effect (nBEKCF) as
\begin{equation}
\label{eq:classifier}
f(\mathbf{X}_r)=\sum_{s=0}^{m-1}\sum_{t=0}^{n-1}\alpha_{s,t} \kappa(\mathbf{X}_r,\mathbf{Z}^{s,t}),
\end{equation}
where $\mathbf{X}_r\in\mathbb{R}^{m\times n\times D}$, and model the ridge regression problem as
\begin{equation}
\label{eq:op-singleimage}
\begin{split}
\min_{\bm{\alpha}} F(\bm{\alpha})\equiv & \sum_{u=0}^{\tilde{M}}\sum_{v=0}^{\tilde{N}}\left (f(\mathbf{X}^{u,v})-y_{u,v}\right)^2 + \lambda \bm{\alpha}^{\top}\bm{\alpha} \\
  = &\left\|\mathbf{K}\bm\alpha-\mathbf{y}\right\|_2^2+\lambda \left \| \bm{\alpha} \right \|_2^2,
\end{split}
\end{equation}
where $\lambda>0$ is the regularization parameter, $\tilde{M}=M-m$, $\tilde{N}=N-n$, $\mathbf{y}= [ y_{0,0},y_{0,1},\ldots,y_{\tilde{M},\tilde{N}-1},y_{\tilde{M},\tilde{N}} ]$ is the vector of gaussian labels, and 
\begin{equation*}
\begin{aligned}
\mathbf{K}=\begin{bmatrix}
\kappa \left ( \mathbf{X}^{0,0},\mathbf{Z}^{0,0} \right ) & \cdots & \kappa \left ( \mathbf{X}^{0,0},\mathbf{Z}^{m-1,n-1} \right ) \\
\vdots & \ddots & \vdots \\
\kappa \left ( \mathbf{X}^{\tilde{M},\tilde{N}},\mathbf{Z}^{0,0} \right ) & \cdots & \kappa \left ( \mathbf{X}^{\tilde{M},\tilde{N}},\mathbf{Z}^{m-1,n-1} \right ) \\
\end{bmatrix}
\end{aligned}
\end{equation*}
is called the \emph{kernel correlation matrix} of $\mathcal{X}$ and $\mathcal{Z}$. Note that the $\bm{\alpha}$ of Problem~\ref{eq:op-singleimage} is not the variable of dual space, thus different from the $\bm{\alpha}$ of KCF.

To solve for $\bm{\alpha^{*}}$, let $\nabla_{\bm\alpha}F(\bm\alpha)=0$; it is achieved that $\left(\mathbf{K^{\top}\mathbf{K}+\lambda\mathbf{I}}\right)\bm\alpha^{*}=\mathbf{K}^{\top}\mathbf{y}$. Because $\mathbf{K}^{\top}\mathbf{K}$ is semi-positive definite, $\mathbf{K}^{\top}\mathbf{K}+\lambda\mathbf{I}$ is invertible. Consequently, the optimal solution of Problem~(\ref{eq:op-singleimage}) is
\begin{equation}
\label{eq:optimal-alpha}
  \bm\alpha^*=\left(\mathbf{K}^{\top}\mathbf{K}+\lambda\mathbf{I}\right)^{-1}\mathbf{K}^{\top}\mathbf{y}.
\end{equation}


The efficiency of calculating $\bm\alpha^*$ is mainly determined by that of constructing $\mathbf{K}$, and the computational burdens of multiplication and inversion of matrices are almost negligible relative to that of constructing $\mathbf{K}$. It is thanks to the real and dense samples in $\mathcal{X}$ and the totally cyclic bases in $\mathcal{Z}$ that the fast construction of $\mathbf{K}$ can be realized. Sec~\ref{sec:fastevaluation} will elaborate how to exploit the density of $\mathcal{X}$ and the cyclicity of $\mathcal{Z}$ to construct $\mathbf{K}$ efficiently without FFT.

It is clear that if $\mathcal{X}$ is cyclic and $\mathcal{Z}\equiv\mathcal{X}$, $\mathbf{K}$ will be a Gram matrix and the \emph{error item} of Problem~(\ref{eq:op-singleimage}) will be exactly the same as that of KCF in Problem~(\ref{eq:kcf-objfunction}). 
But, the learned $f(\cdot;\mathbf{Z}^{s,t})$ in Eq.(\ref{eq:classifier}) is essentially different from Eq.(\ref{eq:kcf-regression}) of KCF, because $\mathcal{Z}\neq\mathcal{X}$, $\mathcal{X}$ is not cyclic and $\mathcal{Z}$ is cyclic in Eq.(\ref{eq:classifier}). It is the characteristics of $\mathcal{Z}$ and $\mathcal{X}$ that make possible the efficient optimization of Problem~(\ref{eq:op-singleimage}) (Sec.~\ref{sec:fastevaluation}). Note that KCF is \emph{not} a special case of nBEKCF because their \emph{regularization items} are different. Sec.~\ref{sec:relatedwork} will present major differences between nBEKCF and the existing CF trackers.

%

\subsection{nBEKCF with Historical Frames}
\label{sec:multipleimages}

In visual tracking task, the appearance model is often trained with multiple frames of different times to improve its robustness. In this section, we will adapt the update scheme proposed in~\cite{dane14a} to our nBEKCF.


Specifically, we model the ridge regression problem of multiple frames as follows.
\begin{equation}
\label{eq:op-multiimage}
\min_{\bm\alpha_Q}F_Q(\bm\alpha_Q)\equiv\sum_{q=1}^{Q}\beta_q\left\|\mathbf{K}_q\bm\alpha_Q-\mathbf{y}\right\|_2^2 + \lambda \left \| \bm{\alpha}_Q \right \|_2^2,
\end{equation}
where $Q$ is the number of historical frames, $q=1$ is the initial frame and $q=Q$ is the present one, $\mathbf{K}_q$ is the kernel correlation matrix of $\mathcal{X}_q$ and $\mathcal{Z}_q$ in frame $q$, $\beta_1=(1-\gamma)^{Q-1}$, $\beta_q=\gamma(1-\gamma)^{Q-q}$ for all $q\geqslant2$, $\sum_{q=1}^{Q}\beta_q=1$, and $\gamma\in[0,1]$ is the learning rate.


Let $\nabla_{\bm\alpha_Q}F_Q(\bm\alpha_Q)=0$. It is achieved that
\[
\label{eq:optimal-alpha-multiimage}
  \bm\alpha_Q^*=\left [\sum_{q=1}^{Q}\beta_q\mathbf{K}_q^{\top}\mathbf{K}_q+\lambda\mathbf{I}\right ]^{-1}\sum_{q=1}^{Q}\beta_q\mathbf{K}_q^{\top}\mathbf{y}.
\]

While the frames come sequentially, an efficient update scheme can be designed as follows.
\begin{subequations}
\begin{align*}
&\bm\alpha_Q^*=(\mathbf{A}_{Q}+\lambda\mathbf{I})^{-1}\mathbf{B}_{Q}, \\
&\mathbf{A}_{Q}=\left ( 1-\gamma \right )\mathbf{A}_{Q-1}+ \gamma \mathbf{K}_Q^{\top}\mathbf{K}_Q, \\
&\mathbf{B}_{Q}=\left ( 1-\gamma \right )\mathbf{B}_{Q-1}+ \gamma \mathbf{K}_Q^{\top}\mathbf{y}, \\
&\hat{\mathbf{X}}_{Q}=\left ( 1-\gamma \right )\hat{\mathbf{X}}_{Q-1}+ \gamma \mathbf{X}_Q, \\
&\hat{\mathbf{Z}}_{Q}=\left ( 1-\gamma \right )\hat{\mathbf{Z}}_{Q-1}+ \gamma \mathbf{Z}_Q,
\end{align*}
\end{subequations}
where $\mathbf{X}_Q$ and $\mathbf{Z}_Q$ are the $\mathbf{X}$ and $\mathbf{Z}$ in frame $Q$, respectively. $\hat{\mathbf{X}}_{Q}$ is updated for the calculation of $\mathbf{K}_Q$. This scheme allows the model to update without storing the previous models. Only the current model $\{ \mathbf{A}_{Q}, \mathbf{B}_{Q}, \hat{\mathbf{X}}_{Q}, \hat{\mathbf{Z}}_{Q} \}$ needs to be saved.




\subsection{Detection of Target Object}
Given the $D$-channel feature map $\mathbf{X}^{\prime}_{Q+1}\in\mathbb{R}^{M\times N\times D}$ of search region in the current frame $Q+1$. Construct $\mathcal{X}^{\prime}_{Q+1}$ and $\hat{\mathcal{Z}}_{Q}$ based on $\mathbf{X}^{\prime}_{Q+1}$ and $\hat{\mathbf{Z}}_{Q}$, respectively, with the methods of constructing $\mathcal{X}$ and $\mathcal{Z}$ presented in Sec.~\ref{sec:hskcfwobes}. Then, the response map of $\mathcal{X}^{\prime}_{Q+1}$ can be obtained with
\[
\mathbf{y}^{\prime}=\mathbf{K}^{\prime}\bm{\alpha}_Q^*,
\]
where $\mathbf{K}^{\prime}$ is the kernel correlation matrix of $\mathcal{X}^{\prime}_{Q+1}$ and $\hat{\mathcal{Z}}_{Q}$.

The element of $\mathbf{y}^{\prime}$ which takes the maximal value is accepted as the optimal location of the target object in frame $Q+1$. The optimal scale of target object is estimated by using DSST~\cite{dane17a}.

\section{Fast Calculation of Correlation Matrix}
\label{sec:fastevaluation}

While solving for the optimal $\bm\alpha^*$ and detecting the target object, kernel correlation matrix, $\mathbf{K}$,\footnote{Refer to $\mathbf{K}$, $\mathbf{K}_Q$, and $\mathbf{K}^{\prime}$ in Sec.~\ref{sec:hskcfwobe}.} has to be constructed first. It is clear that $\mathbf{K}$ can be constructed by using the brute-force approach with computational complexity $O\left ( m^2n^2MND \right )$ because $(\tilde{M}+1)(\tilde{N}+1)\times mn$ elements are contained and the calculation of each element involves two samples of $mnD$ dimensions. This complexity, however, is too high for some time-sensitive tasks such as visual tracking, because $m^2n^2MND$ is often too large there. Typically, $M=N=60$, $m=15$, $n=20$, and $D=31+10$ when HOG~\cite{dalal2005} and CN~\cite{weijer09} are adopted with the cell size being $4\times4$ in CF trackers~\cite{tangm15,dane17b,tangm18}. Therefore, it is necessary to develop a fast algorithm to construct $\mathbf{K}$. Otherwise, it is almost impossible to apply nBEKCF in such time-sensitive tasks. On the other hand, it is noticed that, while solving for $\bm\alpha^*$, the computational cost of matrix inversion is usually not a main bottleneck for efficient solution if the inversion is achieved through solving a system of linear equations, because $\mathbf{K}^{\top}\mathbf{K}\in\mathbb{R}^{mn\times mn}$ and $mn$ is usually not too large, \ie, 300 in the above example.


While constructing $\mathbf{K}$, most common kernels, such as dot-product kernels, polynomial kernels, and Gaussian kernel~\cite{henriques15}, can be employed to calculate its elements. The calculation of these kernels in calculating $\mathbf{K}$ is involved in the autocorrelation of $\mathcal{X}$,
\begin{equation*}
\label{eq:XX}
\begin{aligned}
\mathcal{X}\circ\mathcal{X}=\begin{bmatrix}
\langle \mathbf{X}^{0,0},\mathbf{X}^{0,0} \rangle, & \cdots, & \langle \mathbf{X}^{0,\tilde{N}}, \mathbf{X}^{0,\tilde{N}} \rangle \\
\vdots & \ddots & \vdots \\
\langle \mathbf{X}^{\tilde{M},0}, \mathbf{X}^{\tilde{M},0} \rangle & \cdots & \langle \mathbf{X}^{\tilde{M},\tilde{N}},\mathbf{X}^{\tilde{M},\tilde{N}} \rangle
\end{bmatrix},
\end{aligned}
\end{equation*}
and the cross-correlation (briefly, correlation) of $\mathcal{X}$ and $\mathcal{Z}$,
\begin{equation*}
\begin{aligned}
\mathcal{X}\diamond\mathcal{Z}=\begin{bmatrix}
 \langle \mathbf{X}^{0,0},\mathbf{Z}^{0,0} \rangle & \cdots & \langle \mathbf{X}^{0,0},\mathbf{Z}^{m-1,n-1} \rangle \\
\vdots & \ddots & \vdots \\
 \langle \mathbf{X}^{\tilde{M},\tilde{N}},\mathbf{Z}^{0,0} \rangle & \cdots & \langle \mathbf{X}^{\tilde{M},\tilde{N}},\mathbf{Z}^{m-1,n-1} \rangle
\end{bmatrix},
\end{aligned}
\end{equation*}
where $\langle\cdot,\cdot\rangle$ is the dot product. It is clear that the computational complexities of constructing $\mathcal{X}\circ\mathcal{X}$ and $\mathcal{X}\diamond\mathcal{Z}$ will be $O(mnMND)$ and $O(m^2n^2MND)$, respectively, if the brute-force approach is employed.

A simple idea is to employ FFT to accelerate the construction of $\mathbf{K}$. FFT, however, can only accelerate $\mathcal{X}\diamond\mathcal{Z}$ with the computational complexity $O(mnMND\log{MN})$. This is still not satisfactory for time-sensitive tasks. The reason that FFT is not the optimal choice for our task is that it can only take advantage of the density of $\mathcal{X}$ but is \emph{not} able to exploit the cyclicity of $\mathcal{Z}$. Moreover, FFT is not able to accelerate $\mathcal{X}\circ\mathcal{X}$.

In this section, by means of the principle of integral image~\cite{viola04}, we will present two novel algorithms, autocorrelation with integral image (ACSII) and cyclic correlation with integral matrix (CCIM), to calculate $\mathcal{X}\circ\mathcal{X}$ and $\mathcal{X}\diamond\mathcal{Z}$ efficiently in space domain. The key idea of ACSII and CCIM is to fully exploit the structures of bases and samples, \ie, the bases are cyclic in $\mathcal{Z}$ and samples are densely sampled in $\mathcal{X}$, to eliminate all redundant computations.



\subsection{Autocorrelation with Squared Integral Image}
\label{sec:acii}
Let $\left \{ \mathbf{x}_{p,q}\in\mathbb{R}^{D} \right \}_{p=0,q=0}^{M-1.N-1}$ enumerate all 2D spatial locations of $\mathbf{X}$. If the brute-force approach is employed to calculate $\mathcal{X}\circ\mathcal{X}$, \ie, to calculate all its elements via
\[
\langle\mathbf{X}^{u,v}, \mathbf{X}^{u,v}\rangle = \sum_{p=u}^{u+m-1} \sum_{q=v}^{v+n-1} \left \| \mathbf{x}_{p,q} \right \|_2^2,
\]
where $u=0,\ldots,\tilde{M}$, $v=0,\ldots,\tilde{N}$, there will exist large amounts of redundant calculations because there are high overlaps between neighboring samples of $\mathcal{X}$, as shown in Fig.\ref{fig:illus-nbekcf1}. Specifically, $\left \| \mathbf{x}_{p,q} \right \|_2^2$ and $\left \| \mathbf{x}_{p,q} \right \|_2^2+\left \| \mathbf{x}_{p^{\prime},q^{\prime}} \right \|_2^2$
will be performed multiple times for most $(p,q)$'s and their neighbors $(p^{\prime},q^{\prime})$'s.

To eliminate the redundancy, we propose a novel algorithm, ACSII, to fast calculate the autocorrelation $\mathcal{X}\circ\mathcal{X}$ by means of integral image $\mathbf{I}\in\mathbb{R}^{M\times N}$ with
\[
I_{p,q}=\sum_{i=0}^{p}\sum_{j=0}^{q}\left \| \mathbf{x}_{i,j} \right \|_2^2
\]
as its elements. In this way, any $\langle\mathbf{X}^{u,v}, \mathbf{X}^{u,v}\rangle$ is calculated in a constant time, \ie, three additions of scalars at most, as follows.
\[
\langle\mathbf{X}^{u,v}, \mathbf{X}^{u,v}\rangle = I_{u+m-1,v+n-1}-I_{u,v+n-1}-I_{u+m-1,v}+I_{u,v}.
\]

The detailed technical steps of ACSII are presented in Algorithm 1. It can be seen that ACSII is really similar to the integral image. In fact, ACSII is the integral image except that it acts on the squared vector norm.

%

\begin{algorithm}
\label{alg:ACII}
\caption{\small{Autocorrelation with Square Integral Image (ACSII)}}
\begin{algorithmic}
\item[-]\textbf{Input}: $\mathbf{X}\in\mathbb{R}^{M\times N\times D}$ with $\mathbf{x}_{i,j}\in\mathbb{R}^D$'s as its elements, $m$, and $n$, where $m\leq M$ and $n\leq N$.
\item[-]\textbf{Output}: $\mathcal{X}\circ\mathcal{X}=\mathbf{B}\in\mathbb{R}^{(M-m+1)\times(N-n+1)}$ with $B_{p,q}$'s as its elements.
\item[-] \textbf{Construct squared image}

Set $M\times N$ matrix $\mathbf{A}$ with $A_{i,j}$'s as its elements,
\begin{itemize}
\item[] $A_{i,j}=\left\|\mathbf{x}_{i,j}\right\|_2^2$.
\end{itemize}
\item[-] \textbf{Construct squared integral image $\mathbf{I}$}
\begin{itemize}
\item[1.] $I_{0,0}=A_{0,0}$.
\item[2.] for $p=1$ to $M-1$:\;\;$I_{p,0}=I_{p-1,0}+A_{p,0}$.
\item[3.] for $q=1$ to $N-1$:\;\;$I_{0,q}=I_{0,q-1}+A_{0,q}$.
\item[4.] for $p=1$ to $M-1$,\;\;$q=1$ to $N-1$:

$I_{p,q}=I_{p-1,q}+I_{p,q-1}-I_{p-1,q-1}+A_{p,q}$.
\end{itemize}
\item[-] \textbf{Calculate autocorrelation $\mathcal{X}\circ\mathcal{X}$}
\begin{itemize}
\item[1.] $B_{0,0}=I_{m-1,n-1}$.
\item[2.] for $p=1$ to $M-m$:

$B_{p,0}=I_{p+m-1,n-1}-I_{p-1,n-1}$.
\item[3.] for $q=1$ to $N-n$:

$B_{0,q}=I_{m-1,q+n-1}-I_{m-1,q-1}$.
\item[4.] for $p=1$ to $M-m$,\;\;$q=1$ to $N-n$:

$B_{p,q}=I_{p+m-1,q+n-1}-I_{p-1,q+n-1}-I_{p+m-1,q-1}+I_{p-1,q-1}$.
\end{itemize}
\end{algorithmic}
\end{algorithm}


\subsection{Cyclic Correlation with Integral Matrix}
\label{sec:ccim}

\begin{figure}[t]
\centering
\includegraphics[height=29mm,width=82mm]{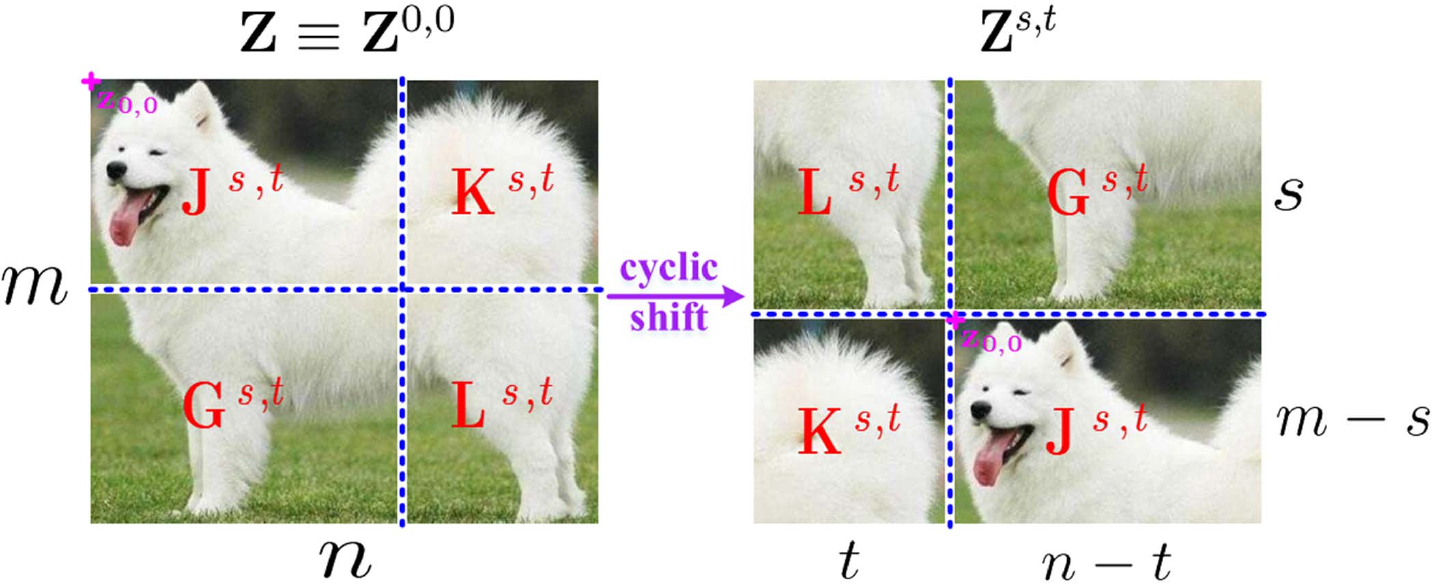}
\caption{The division of cyclic base $\mathbf{Z}^{s,t}$ into four sub-bases, $\mathbf{L}^{s,t}$, $\mathbf{G}^{s,t}$, $\mathbf{K}^{s,t}$, and $\mathbf{J}^{s,t}$, by the location of $\mathbf{z}_{0,0}$. The relative spatial locations of elements of $\mathbf{D}^{s,t}$ are the same as their relative locations in $\mathbf{Z}$, where $\mathbf{D}^{s,t}\in\{\mathbf{L}^{s,t},\mathbf{G}^{s,t},\mathbf{K}^{s,t},\mathbf{J}^{s,t}\}$.
}
\label{fig:divisionofcf}
\vspace{-4mm}
\end{figure}

Suppose $\mathbf{H}_{((a_1,b_1),(a_2,b_2))}$ is the sub-matrix of matrix $\mathbf{H}$ with $(a_1,b_1)$ and $(a_2,b_2)$ as its top-left and down-right corners. Let $\{\mathbf{z}_{s,t}\in\mathbb{R}^{D}\}_{s=0,t=0}^{m-1,n-1}$ enumerate all 2D spatial locations of $\mathbf{Z}$. If the brute-force approach is employed to calculate $\mathcal{X}\diamond\mathcal{Z}$, \ie, to calculate $\mathcal{X}\diamond\mathcal{Z}$ through
\begin{small}
\begin{equation}
\label{eq:XZ}
\mathcal{X}\diamond\mathcal{Z} =[\text{vec}(\mathbf{Z}^{0,0}\text{\footnotesize\FiveStarOpen}\mathbf{X}), \cdots, \text{vec}(\mathbf{Z}^{m-1,n-1}\text{\footnotesize\FiveStarOpen}\mathbf{X})],
\end{equation}
\end{small}
where \text\footnotesize{\FiveStarOpen} is the correlation operator and $\text{vec}(\mathbf{H})$ indicates the vectorization of $\mathbf{H}$, there will be large amounts of redundant calculations because all bases, $\mathbf{Z}^{s,t}$'s, of $\mathcal{Z}$ are obtained by cyclically shifting $\mathbf{Z}$. In fact, $\langle\mathbf{z}_{s,t}, \mathbf{x}_{p,q}\rangle$ and $\langle\mathbf{z}_{s,t}, \mathbf{x}_{p,q}\rangle+\langle\mathbf{z}_{s+\Delta s,t+\Delta t}, \mathbf{x}_{p+\Delta s,q+\Delta t}\rangle$ will be performed multiple times for most $s$, $t$, $p$, and $q$.

As shown in Fig.~\ref{fig:divisionofcf}, according to the cyclicity of $\mathcal{Z}$, any base $\mathbf{Z}^{s,t}\in\mathcal{Z}$ can always be divided into four sub-bases, $\mathbf{L}^{s,t}$, $\mathbf{G}^{s,t}$, $\mathbf{K}^{s,t}$, and $\mathbf{J}^{s,t}$, according to the location of $\mathbf{z}_{0,0}$, where $\mathbf{z}_{0,0}$ is the top-left element of $\mathbf{Z}$, and the relative spatial locations of elements of each sub-base are the same as their relative locations in $\mathbf{Z}$. Therefore, $\mathbf{Z}^{s,t}\text{\footnotesize\FiveStarOpen}\mathbf{X}$ can be decomposed into four items,
\begin{equation*}
\begin{aligned}
\mathbf{Z}^{s,t}\text{\footnotesize\FiveStarOpen}\mathbf{X} &=(\mathbf{L}^{s,t}\text{\footnotesize\FiveStarOpen}\mathbf{X})_{((0,0),(M-m,N-n))}\\
&+(\mathbf{G}^{s,t}\text{\footnotesize\FiveStarOpen}\mathbf{X})_{((0,t),(M-m,N-n+t))}\\
&+(\mathbf{K}^{s,t}\text{\footnotesize\FiveStarOpen}\mathbf{X})_{((s,0),(M-m+s,N-n))}\\
&+(\mathbf{J}^{s,t}\text{\footnotesize\FiveStarOpen}\mathbf{X})_{((s,t),(M-m+s,N-n+t))}.
\end{aligned}
\end{equation*}

If $\langle\mathbf{z}_{s,t}, \mathbf{x}_{p,q}\rangle$'s for all possible $(s,t)$'s and $(p,q)$'s are calculated and stored, then $\mathbf{D}^{s,t}\text{\footnotesize\FiveStarOpen}\mathbf{X}$, where $\mathbf{D}^{s,t}\in\{\mathbf{L}^{s,t},\mathbf{G}^{s,t},\mathbf{K}^{s,t},\mathbf{J}^{s,t}\}$, can be obtained by first retrieving stored calculations and then summing them. In this way, the multiple calculations on $\langle\mathbf{z}_{s,t}, \mathbf{x}_{p,q}\rangle$'s are eliminated. In our algorithm, $\langle\mathbf{z}_{s,t}, \mathbf{x}_{p,q}\rangle$'s for all possible $(s,t)$'s and $(p,q)$'s are called fundamental calculation, and stored as a series of fundamental matrices. In order to obtain $\mathbf{Z}^{0,0}\text{\footnotesize\FiveStarOpen}\mathbf{X}$ by the summation of all fundamental matrices, cyclic shifting is performed on the fundamental matrices. Each element of $\mathbf{Z}^{0,0}\text{\footnotesize\FiveStarOpen}\mathbf{X}$ equals to an element of the summation.

Another part of repeated calculations is the summation. It is clear that large amounts of summations calculated in $\mathbf{D}^{s,t}\text{\footnotesize\FiveStarOpen}\mathbf{X}$ are repeated in $\mathbf{D}^{s^{\prime},t^{\prime}}\text{\footnotesize\FiveStarOpen}\mathbf{X}$, where $s\neq s^{\prime}$ or $t\neq t^{\prime}$. For example, if $s^{\prime}=s+1$ and $t^{\prime}=t$, most summations of fundamental calculations involved in $\mathbf{J}^{s,t}\text{\footnotesize\FiveStarOpen}\mathbf{X}$ are repeated by those involved in $\mathbf{J}^{s^{\prime},t^{\prime}}\text{\footnotesize\FiveStarOpen}\mathbf{X}$. Because the fundamental calculations are stored as the fundamental matrices in our algorithm, the repeated summations are the repeated summations of fundamental matrices. To eliminate all these repeated summations, inspired by the integral image, we design the integral matrix $\mathbf{M}$ so as to calculate each $\mathbf{D}^{s,t}\text{\footnotesize\FiveStarOpen}\mathbf{X}$ in a constant time. Because the fundamental matrices are constructed according to the calculation of $\mathbf{Z}^{0,0}\text{\footnotesize\FiveStarOpen}\mathbf{X}$, $\mathbf{D}^{s,t}\text{\footnotesize\FiveStarOpen}\mathbf{X}$'s have to be cyclically shifted to align each other and then are summed to achieve $\mathbf{Z}^{s,t}\text{\footnotesize\FiveStarOpen}\mathbf{X}$ correctly, if $s\neq 0$ or $t\neq 0$.

Consequently, $\mathbf{Z}^{s,t}\text{\footnotesize\FiveStarOpen}\mathbf{X}$ is calculated efficiently. Note that $\mathbf{M}$ is a block matrix and the same as the integral image in principle. The difference between them is that $\mathbf{M}$ operates on block matrices, whereas the integral image on scalars.


According to the above analysis, we develop the algorithm, cyclic correlation with integral matrix (CCIM), to calculate $\mathcal{X}\diamond\mathcal{Z}$ efficiently. The formal steps of CCIM is presented in Algorithm 2. The appendices provide an example to illustrate how CCIM works exactly and the proof of its correctness.

\begin{algorithm}[t]
\caption{\small{Cyclic Correlation with Integral Matrix (CCIM)}}
\begin{algorithmic}
\item[-]\textbf{Input}: $\mathbf{Z}\in\mathbb{R}^{m\times n\times D}$ with $\mathbf{z}_{s,t}\in\mathbb{R}^D$'s as its elements, $\mathbf{X}\in\mathbb{R}^{M\times N\times D}$, $m\leqslant M$, and $n\leqslant N$.
\item[-]\textbf{Output}: \\
    $\mathcal{X}\diamond\mathcal{Z} =[\text{vec}(\mathbf{Z}^{0,0}\text{\footnotesize\FiveStarOpen}\mathbf{X}),\cdots, \text{vec}(\mathbf{Z}^{m-1,n-1}\text{\footnotesize\FiveStarOpen}\mathbf{X})]$.
\item[-] \textbf{Construct fundamental matrices}

\item[] /* Cyclically shift fundamental calculations $\mathbf{z}_{s,t}\text{\footnotesize\FiveStarOpen}\mathbf{X}$'s to construct fundamental matrices $\mathbf{B}^{s,t}$'s so as to $\sum_{s,t}\mathbf{B}^{s,t}_{((0,0),(M-m,N-n))}=\mathbf{Z}\text{\footnotesize\FiveStarOpen}\mathbf{X}$.

\begin{itemize}
\item[] for $s=0$ to $m-1$,\;\;$t=0$ to $n-1$:
\begin{itemize}
\item[] 
    $\mathbf{B}^{s,t}=\mathbf{P}^{-s}_M\left(\mathbf{z}_{s,t}\text{\footnotesize\FiveStarOpen}\mathbf{X}\right)
    \mathbf{Q}^{-t}_N$.
\end{itemize}
\end{itemize}
\item[-] \textbf{Construct integral matrix} $\mathbf{M}\equiv\left[\mathbf{M}^{s,t}\right]_{m\times n}$ with $\mathbf{M}^{s,t}\in\mathbb{R}^{M\times N}$ being its block element.
\begin{itemize}
\item[1.] $\mathbf{M}^{0,0}=\mathbf{B}^{0,0}$,
\item[2.] for $s=1$ to $m-1$:\;\;$\mathbf{M}^{s,0}=\mathbf{M}^{s-1,0}+\mathbf{B}^{s,0}$.
\item[3.] for $t=1$ to $n-1$:\;\;$\mathbf{M}^{0,t}=\mathbf{M}^{0,t-1}+\mathbf{B}^{0,t}$.
\item[4.] for $s=1$ to $m-1$,\;\;$t=1$ to $n-1$:

$\mathbf{M}^{s,t}=\mathbf{M}^{s-1,t}+\mathbf{M}^{s,t-1}-\mathbf{M}^{s-1,t-1}+\mathbf{B}^{s,t}$.
\end{itemize}
\item[-] \textbf{Calculate correlation} $\mathbf{Z}^{s,t}\text{\footnotesize\FiveStarOpen}\mathbf{X}$.
\begin{itemize}
\item[] for $s=0$ to $m-1$,\;\;$t=0$ to $n-1$:

\hskip -4mm /* Divide $\mathbf{Z}^{s,t}\text{\footnotesize\FiveStarOpen}\mathbf{X}$ into four parts and calculate each of them one by one by means of $\mathbf{M}$.
\end{itemize}
\begin{itemize}
\item[1.] $\mathbf{L} = \mathbf{M}^{m-1,n-1}-\mathbf{M}^{m-1,n-t-1}-\mathbf{M}^{m-s-1,n-1}+\mathbf{M}^{m-s-1,n-t-1}$.

    If $s=0$ or $t=0$, $\mathbf{L}=\mathbf{0}_{M\times N}$.
\item[2.] $\mathbf{G}=\mathbf{M}^{m-1,n-t-1}-\mathbf{M}^{m-s-1,n-t-1}$.

    If $s=0$, $\mathbf{G}=\mathbf{0}_{M\times N}$.
\item[3.] $\mathbf{K}=\mathbf{M}^{m-s-1,n-1}-\mathbf{M}^{m-s-1,n-t-1}$.

    If $t=0$, $\mathbf{K}=\mathbf{0}_{M\times N}$.
\item[4.] $\mathbf{J}=\mathbf{M}^{m-s-1,n-t-1}$.

\hskip -4mm /* Align the results of four sub-correlations:
\item[5.] $\mathbf{L}\leftarrow\mathbf{P}_M^{m-s}\mathbf{L}\mathbf{Q}_N^{n-t}$,
\item[6.] $\mathbf{G}\leftarrow\mathbf{P}_M^{m-s}\mathbf{G}\mathbf{Q}_N^{-t}$,
\item[7.] $\mathbf{K}\leftarrow\mathbf{P}_M^{-s}\mathbf{K}\mathbf{Q}_N^{n-t}$,
\item[8.] $\mathbf{J}\leftarrow\mathbf{P}_M^{-s}\mathbf{J}\mathbf{Q}_N^{-t}$.

\hskip -4mm /* Sum the aligned four sub-correlations: 
\item[9.] $\mathbf{Z}^{s,t}\text{\footnotesize\FiveStarOpen}\mathbf{X}=(\mathbf{L}+\mathbf{G}+
\mathbf{K}+\mathbf{J})_{((0,0),(M-m,N-n))}$.
\end{itemize}
\end{algorithmic}
\end{algorithm}

\subsection{Characteristics of ACSII and CCIM}
\label{sec:compcomlexity}

\begin{table*}
\begin{center}
\resizebox{1\textwidth}{!}{
\begin{tabular}{|c||c|c|c|c|c|c|}\hline
      & Object of   & Computational & Elapsed Time & Space or & Exploiting & Exploiting \\
      & Calculation & Complexity    &  ($M=N=60$, $m=15$, $n=20$, $D=41$) & Frequency Domains & Density of $\mathcal{X}$ & Cyclicity of $\mathcal{Z}$ \\ \hline\hline
ACSII & $\mathcal{X}\circ\mathcal{X}$ & $O(MND)$ & $3ms$ & Space & Yes & - \\ \hline
CCIM  & $\mathcal{X}\diamond\mathcal{Z}$ & $O(mnMND)$ & $20ms$ & Space & Yes & Yes \\ \hline
FFT   & $\mathcal{X}\diamond\mathcal{Z}$ & $O(mnMND\log MN)$ & $530ms$ & Frequency & Yes & No \\ \hline
\end{tabular}
}
\end{center}
\caption{Characteristics of ACSII, CCIM, and FFT in our task. `-' means the algorithm is not involved in the data. See Sec.~\ref{sec:compcomlexity} for details.}
\label{tab:compcomplexity}
\end{table*}

The computational complexity of ACSII is $O(MND)$ because those of its three steps, constructing squared image, constructing squared integral image, and calculating autocorrelation, are $O(MND)$, $O(MN)$, and $O(MN)$, respectively.

Because the computational complexities of the three steps of CCIM, construct fundamental matrices, construct integral matrix, and calculate correlation, are $O(mnMND)$, $O(mnMN)$, and $O(mnMN)$, respectively, its computational complexity is $O(mnMND)$, much lower than that with FFT.

In order to experimentally verify the superior efficiency of CCIM over that of FFT, we calculate $\mathcal{X}\diamond\mathcal{Z}$ on a PC with Intel Core i7 CPU under the typical situation stated in the beginning of Sec.~\ref{sec:fastevaluation}. It is not surprising to see that CCIM takes only 20 ms, while FFT 530 ms, to achieve $\mathcal{X}\diamond\mathcal{Z}$. In addition, ACSII takes 3 ms to calculate $\mathcal{X}\circ\mathcal{X}$.

By fully exploiting the structures of bases and samples, ACSII and CCIM eliminate all redundant computations in calculating $\mathcal{X}\circ\mathcal{X}$ and $\mathcal{X}\diamond\mathcal{Z}$ in space domain. It is seen from the above analysis that the efficiency of CCIM is much higher than that of FFT in both theory and practice, because FFT cannot use the relations among the bases. Table~\ref{tab:compcomplexity} summarizes the characteristics of ACSII, CCIM, and FFT in our task.

\section{Related Work}
\label{sec:relatedwork}
Our novel nBEKCF is essentially different from all existing CF trackers, because it is NOT based on KCF at all. Specifically, there are three key differences between nBEKCF and existing CF trackers. 1) The theoretical foundations are different. 2) The sampling strategies are different (non-cyclic vs. cyclic). And 3) the efficient optimization procedures are different (non-FFT vs. FFT-related). The root reason to cause these differences is the separation of the sets of real and dense training samples and cyclic bases in our nBEKCF. In contrast, the both sets are identical and cyclic in KCF.

We do not model our problem with Tikhonov regularization~\cite{tangm18} when deriving nBEKCF. Therefore, Problem~(\ref{eq:op-singleimage}) does not have to be transferred into the dual space by means of Representer Theorem~\cite{scholkopf2002} to solve.

In nBEKCF, we also do not require the kernelized correlation filter $f$ to lie in a bounded convex subset of a reproducing kernel Hilbert space (RKHS) defined by a positive definite kernel function $\kappa(\cdot,\cdot)$. $S_{\kappa}\equiv\{\kappa(\cdot,\mathbf{X}^{u,v})\}$ can be considered as a set of basis functions, and it will be orthogonal if $f$ lies in a bounded convex subset of a RKHS defined by $\kappa(\cdot,\cdot)$.

In almost all existing CF trackers, the training sets are constructed through cyclically shifting a single image patch (\ie, base sample) in a pixel-wise way, and the training samples are the cyclically shifted base samples themselves (e.g., in KCF, MKCF~\cite{tangm15,tangm18}, and SRDCF) or their clipped parts (e.g., in BACF). Therefore, there are more or less synthetic samples in their training sets. In our nBEKCF, however, the training samples are drawn from an image region in the traditional sliding window way, and there is no cyclic shifting of the image region. Therefore, all training samples are non-synthetic, \ie, real, in nBEKCF, avoiding the boundary effect completely.

Unlike almost all existing CF trackers which resort to FFT for fast optimization, we develop two novel algorithms, ACSII and CCIM, to achieve efficient training and detection. CCIM is even more efficient than its FFT counterpart in nBEKCF.

Our previous work, fdKCF*~\cite{fdkcf-zhengly19}, optimizes the regression model of KCF in space domain, eliminates almost all repeated calculations, and then uses GPU to accelerate the remaining ones, avoiding boundary effects of KCF. It is clear that nBEKCF is different from fdKCF* because their regression models and optimization schemes are both totally different. While nBEKCF applies ACSII and CCIM to accelerate its optimization, fdKCF* relies on the look-up table to efficiently optimize. On the other hand, the major computational costs of fdKCF* and nBEKCF are $O(M^2N^2(D+mn))$ and $O(MNDmn)$, respectively. Because $M>m$ and $N>n$, the cost of fdKCF* is much larger than that of nBEKCF. In practice, only on GPU can fdKCF* run in super-real-time even if low dimensional hand-crafted features are employed, because one of its major computational costs, $O(M^2N^2mn)$, is large and irrelative to the dimensionality of features. Whereas, nBEKCF can construct its kernel correlation matrices in a high efficiency, running in super-real-time on CPU if low dimensional hand-crafted features are employed. Note that the speed of constructing a kernel correlation matrix of nBEKCF on CPU is really close to that of fkKCF* on GPU, if the pre-trained network features of high dimensionality are exploited.

In comparison to Siamese trackers~\cite{SiamFC-berti16b,SiamRPN-Lib18,SiamDW-zhangz19,SiamRPN++-lib19,SiamGCN-gaoj19}, our nBEKCF, as a CF tracker, can be trained discriminatively, while they cannot, as pointed out by~\cite{DiMP-bhat19}. That is, nBEKCF utilizes both foreground and background to train its filter, while Siamese trackers only use the target appearance when training their models. Consequently, in terms of trackers themselves, nBEKCF is more robust than Siamese trackers for various backgrounds.

\section{Experiments}
\label{sec:experiments}

We implement two versions of our nBEKCF, nBEKCF-HC and nBEKCF-D by employing hand-crafted features and deep convolutional neural networks (CNNs) features, respectively. nBEKCF-HC is evaluated on three public benchmarks, OTB-2013, OTB-2015, and NfS, and compared against state-of-the-art trackers with hand-crafted features. nBEKCF-D is evaluated on other three public benchmarks, VOT2018, GOT10k, and TrackingNet, and compared to state-of-the-art trackers with CNNs features. We do not test nBEKCF-HC on the latter three ones because the hand-crafted features are too weak on them.

In our experiments, for a fair comparison, the same type of state-of-the-arts trackers are among the list of compared trackers. The same type of state-of-the-arts trackers means the top ones with the similar motivation, similar features, and similar scale adaptation scheme to the nBEKCF's. Therefore, ECO-HC~\cite{dane17b} is selected from the trackers with hand-crafted features, and UPDT~\cite{bhat18} from the trackers with ImageNet pre-trained features, which, like nBEKCF, both adopt the scale-pyramid scheme to decide proper scales of target.


\subsection{Implementation Details}

In nBEKCF-HC, as in the top CF tracker with hand-crafted features, ECO-HC, HOG with $31$-channels and color-name (CN) with $10$-channels are employed, and both their cell sizes are $4\times 4$. Gaussian kernel is applied with standard deviation $6$ to construct the kernel correlation matrices. The learning rate $\gamma$ is $0.008$. The size of learning or search region is $M=N=3\sqrt{mn}$ for the tradeoff between localization accuracy and speed. ACSII and CCIM are implemented in C++, and other parts in Matlab. The experiments are performed on a single Intel I7 CPU.

In nBEKCF-D, as in the top CF tracker with ImageNet pre-trained features, UPDT, the Conv-1 and Block-4 feature maps of ResNet50 are exploited, and the bilinear interpolation is applied to increase the resolution of the Block-4 feature maps $4\times 4$ times. Linear kernel is used to construct the kernel correlation matrices. The learning rate $\gamma=0.004$. The size of learning or search region is $M=N=4.5\sqrt{mn}$. ACSII and CCIM are implemented in C++ with CUDA, and other parts in Pytorch. The experiments are performed on a single NVIDIA TITAN X GPU.

Regularization parameter $\lambda=0.01$. Gaussian response $\mathbf{y}$ is identical to that in KCF with variance $0.01$.

\subsection{Ablation Study}
\label{sec:ablation}

Our nBEKCF is the first CF tracker\footnote{The original version of this paper (arXiv:1806.06406) was posted on arXiv.org in June, 2018, much earlier than fdKCF*~\cite{fdkcf-zhengly19}} which is not plagued by the boundary effect at all and is able to exploit non-linear kernels and high-dimensional hand-crafted features, while running in a high efficiency with single CPU. In this section, we investigate the impacts of avoiding the boundary effect thoroughly and choosing different kernels in nBEKCF. We conduct the ablation experiments with nBEKCF-HC on OTB2015~\cite{wu15}.

As a baseline tracker in our ablation study, SAMF~\cite{liy14} is the KCF with scale-pyramid scheme, exploits HOG and CN as its features, and suffers from the boundary effect. Because the motivation of BACF and SRDCF is also to address the boundary effect of KCF and they both employ scale-pyramid scheme to determine the scale of target, they are compared with SAMF and nBEKCF in this section. For the fair comparison, we implement SRDCF-HC and BACF-HC by adding color features (CN) into the public source codes of SRDCF and BACH, and the linear kernel is employed in nBKECF-HC because SRDCF and BACF are not able to employ any non-linear kernel.



Table~\ref{tab:ablation} shows the results where AUC~\cite{wu15} is used to evaluate their accuracies. It is seen from the table that the right four trackers outperform the base tracker SAMF-HC with large margins, and nBEKCF-HC-L performs better than SRDCF-HC and BACF-HC. Therefore, alleviating or avoiding the boundary effect will increase the accuracy, and avoiding the boundary effect completely in nBEKCF improves the accuracy more than just alleviating it in SRDCF-HC and BACF-HC.

On the other hand, the accuracy of nBEKCF-HC is higher than that of nBEKCF-HC-L, confirming that employing non-linear kernels will achieve a higher accuracy than employing the linear one, as shown by Zuo~\etal~\cite{zuo-pami19}.


Table~\ref{tab:ablation} also shows that nBEKCF-HC-L and nBEKCF-HC not only outperform other trackers, but are able to run at much faster speeds as well. It is interesting to notice that nBEKCF-HC-L and nBEKCF-HC even run faster than SAMP which, exactly like KCF, is plagued by the boundary effect. Note that our estimation of FPS's contains all steps of nBEKCF (feature extraction, model training and updating, ACSII, CCIM, and localization), except for reading the frames,  according to the common way in visual tracking domain.

\begin{table}
\begin{center}
\resizebox{1\columnwidth}{!}{
\begin{tabular}{|r||c|c|c|c|c|}\hline
     & SAMF  & SRDCF-HC & BACF-HC & nBEKCF-HC-L & nBEKCF-HC \\ \hline\hline
AUC  & 0.572 & 0.616    & 0.620   & 0.632       & 0.643     \\ \hline
mFPS & 40    & 3        & 30      & 55          & 50        \\ \hline
\end{tabular}}
\end{center}
\caption{AUCs and mean FPSs on OTB-2015. ``-HC'' means the indicated tracker exploits HOG and CN features, and ``-L'' indicates the tracker applies the linear kernel. See Sec.~\ref{sec:ablation} for details.}
\label{tab:ablation}
\end{table}

\subsection{Evaluation on OTB-2013, OTB-2015, and NfS}
\label{sec:evaotbnfs}

\begin{figure*}
\begin{center}
\subfloat[]{
\label{fig:otb.13}
\includegraphics[width=0.31\textwidth]{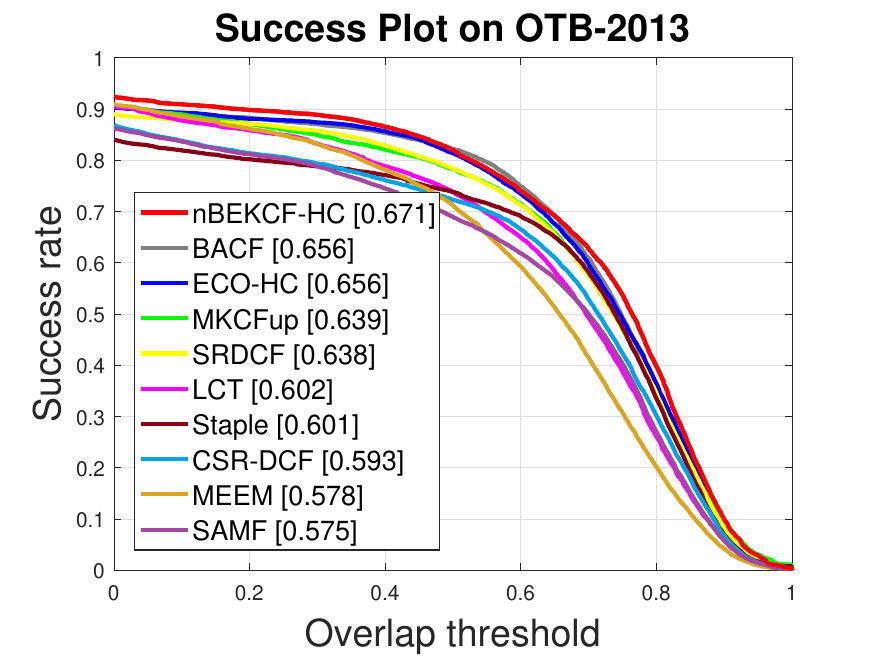}}
\subfloat[]{
\label{fig:otb.15}
\includegraphics[width=0.31\textwidth]{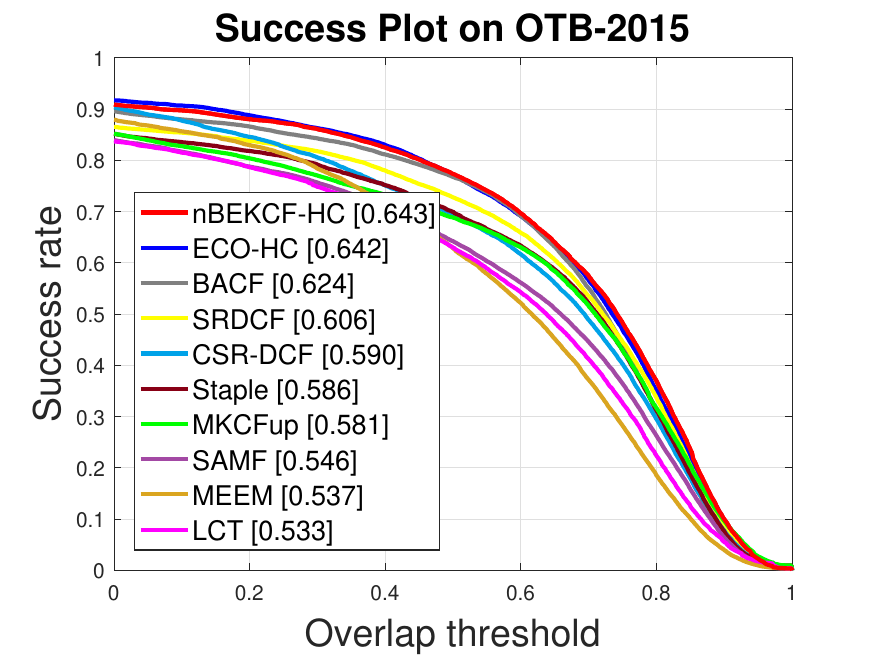}}
\subfloat[]{
\label{fig:nfs}
\includegraphics[width=0.31\textwidth]{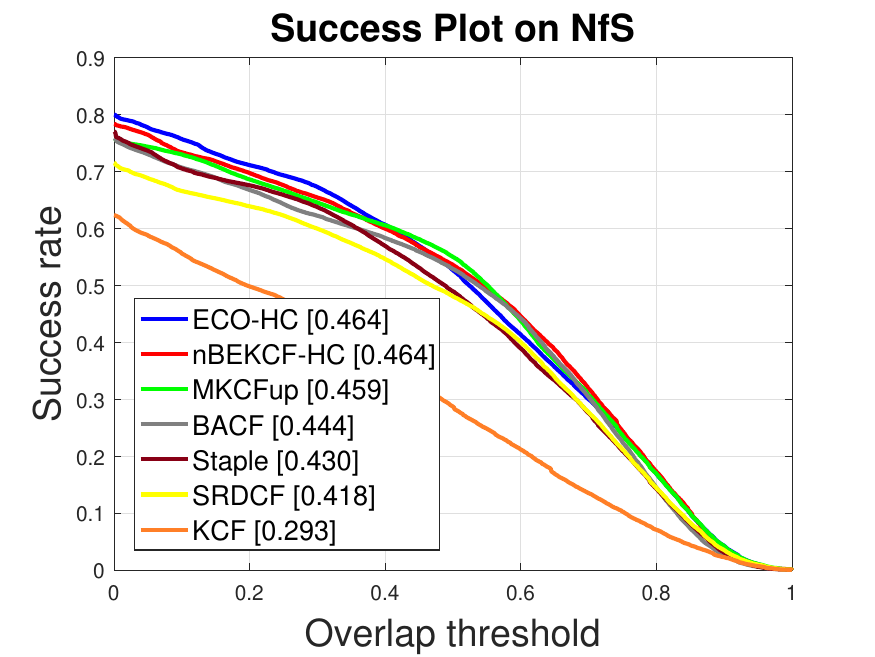}}
\end{center}
\caption{The mean success plots of our nBEKCF-HC and other state-of-the-art trackers with hand-crafted features on OTB-2013, OTB-2015, and NfS. The mean AUCs are reported in the legend. nBEKCF-HC achieves the best results.}
\label{fig:otb}
\end{figure*}


\begin{table}
\begin{center}
\resizebox{0.98\columnwidth}{!}{
\begin{tabular}{|c|c|c|c|c|c|}
\hline
Tracker &      nBEKCF-HC    & BACF  & ECO-HC & LCT &       MKCFup  \\ \hline
  mFPS  & 50 &  35   &   40   &  30 & 150 \\ \hline \hline
Tracker & CSR-DCF & SRDCF & MEEM &        Staple      & SAMF  \\ \hline
  mFPS  &    15   &  10   &  15  & 70 &  30    \\ \hline	
\end{tabular}}
\end{center}
\caption{The mean FPSs of our nBEKCF-HC and other state-of-the-art trackers with hand-crafted features.}
\label{table:fps}
\end{table}

{\noindent \bfseries OTB-2013/2015}~\cite{wu13,wu15}. OTB-2013 and OTB-2015 are the most popular benchmarks with various challenges for the evaluation of trackers, and contain 50 and 100 videos, respectively. On the OTB-2013 and OTB-2015 experiments, we compare our nBEKCF-HC against nine hand-crafted feature based state-of-the-art trackers, MEEM~\cite{zhangj14}, SAMF~\cite{liy14}, SRDCF~\cite{dane15a}, LCT~\cite{mac15a}, Staple~\cite{bertinetto16}, BACF~\cite{galoo17a}, ECO-HC~\cite{dane17b}, CSR-DCF~\cite{luke17}, and MKCFup~\cite{tangm18}. All trackers are evaluated by success plot and AUC. Figs.~\ref{fig:otb.13} and \ref{fig:otb.15} show that our nBEKCF-HC obtains the mean AUC of $67.1\%$ and $64.3\%$ on OTB-2013 and OTB-2015, respectively, outperforming all other state-of-the-art CF trackers with hand-crafted features.

In addition, table~\ref{table:fps}~\footnote{We ran nBEKCF-HC, BACF, and ECO-HC on an identical PC for the fair comparison of fps.} shows the mean FPS of the above trackers on OTB-2015. It can be seen that the running speed of our nBEKCF-HC is faster than all other trackers, except for MKCFup and Staple which take full advantage of the cyclic structure of samples. Although MKCFup and Staple are two fastest trackers, they suffer from the boundary effect, resulting in lower localization accuracies. It is worth noting that ECO-HC applies several tricks, such as sparse update and feature dimension reduction, to speed up. Whereas, nBEKCF-HC does not employ any similar tricks but is still faster than ECO-HC.

\vskip 2mm
{\noindent \bfseries NFS}~\cite{galoo17b}. We evaluate nBEKCF-HC on the 240 FPS version of NfS benchmark which contains 100 challenging videos. On the NfS experiment, we compare nBEKCF-HC against five representative CF trackers, KCF, SRDCF, ECO-HC, BACF, and MKCFup. All trackers are evaluated by success plot and AUC. Fig.~\ref{fig:nfs} shows that our nBEKCF-HC obtains the mean AUC of $0.464$, outperforming all other CF trackers, except for ECO-HC. It is worth noting that ECO-HC uses sample clustering to improve its accuracy. Whereas, nBEKCF-HC does not apply any similar tricks but still achieves competitive accuracy with ECO-HC. In addition, it is seen that the success rate at overlap threshold 0.5, \ie, the mean overlap precision, of nBEKCF-HC is slightly higher than that of ECO-HC.

It can be concluded from the above experiments that our nBEKCF-HC achieves the best trade-off between accuracy and speed among all hand-crafted feature based trackers.
\subsection{Evaluation on VOT, GOT10k, and TrackingNet}
\label{sec:vot}

\begin{figure}
\centering
  \includegraphics[width=0.41\textwidth]{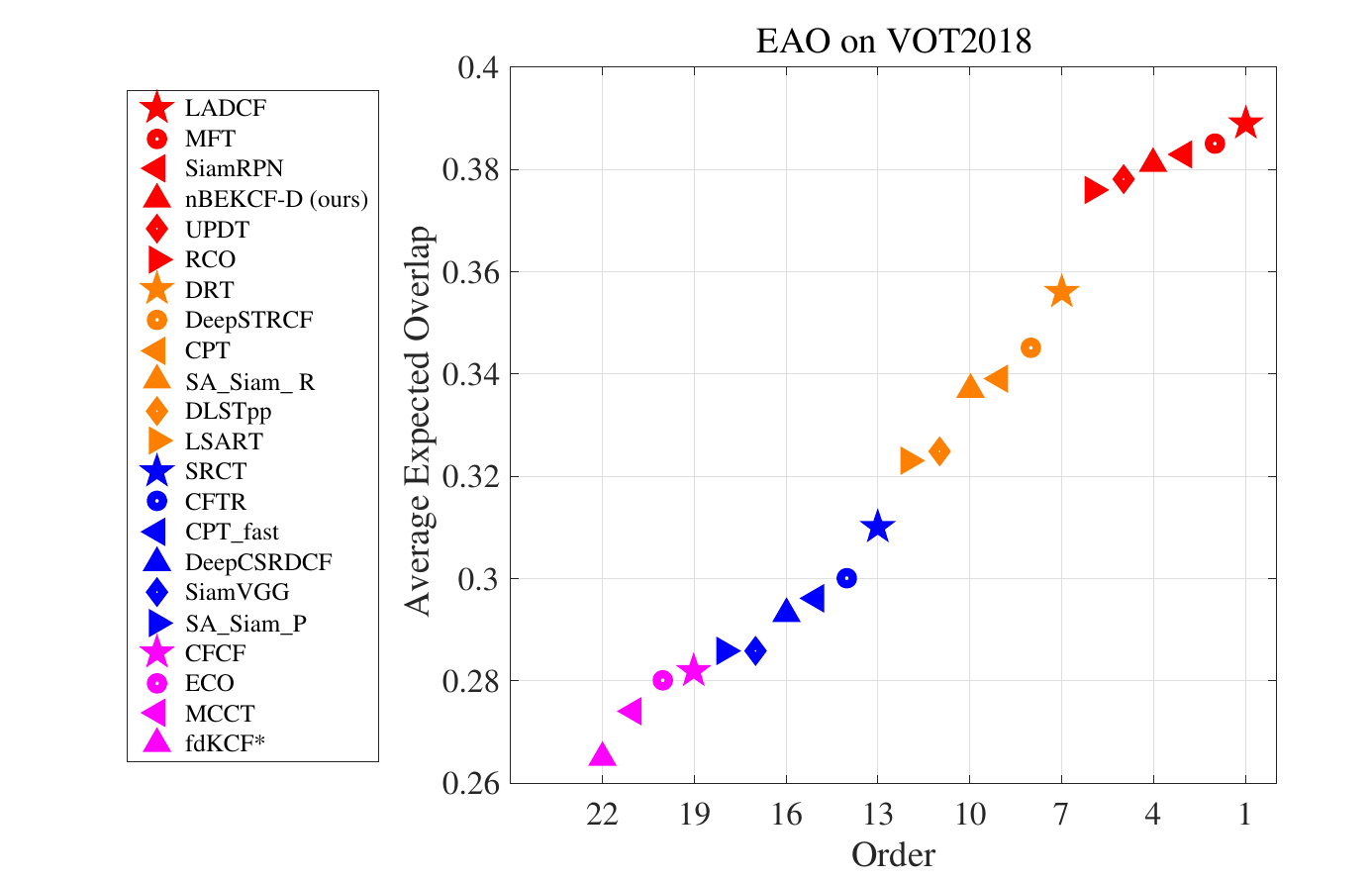}
  \caption{Comparison among nBEKCF-D and state-of-the-art trackers on VOT2018 challenge in terms of EAO.}
\label{fig:vot18-eao}
\end{figure}

{\noindent \bfseries VOT2018}~\cite{vot18}. We evaluate our nBEKCF-D on Visual Object Tracking (VOT) 2018 challenge which consists of 60 sequences. On the VOT2018 experiment, we compare nBEKCF-D against the latest CF tracker fdKCF*~\cite{fdkcf-zhengly19} along with the top-20 trackers on VOT2018 challenge. Following the VOT challenge protocol, all trackers are evaluated by expected average overlap (EAO). Fig.~\ref{fig:vot18-eao} shows the result. It is seen from the figure that our nBEKCF-D achieves EAO $0.381$, outperforming most state-of-the-art CF trackers, CCOT~\cite{dane16b}, ECO~\cite{dane17b}, CFCF~\cite{gundogdu18}, UPDT~\cite{bhat18}, and fdKCF*. In fact, the main reason that SiamRPN outperforms nBEKCF-D is that the former trains a deep network to regress the bounding boxes of targets finely on largest datasets, while the latter only employs a scale-pyramid scheme.

\begin{table}
\begin{center}
\resizebox{1\columnwidth}{!}{
\begin{tabular}{|r|c|c|c|c|c|c|c|c|}
\toprule
Tracker          &        nBEKCF-D         &  HCF  & ECO  & CCOT & MDNet &        SiamFC          &          CFNet          &        GOTURN\\
\midrule
SR$_{0.50}$ (\%) & \textcolor{red}{44.8}   &  29.7 & 30.9 & 32.8 &  30.3 &        35.3            & \textcolor{green}{40.4} & \textcolor{blue}{37.5}\\
\midrule
SR$_{0.75}$ (\%) & \textcolor{green}{12.5} &  8.8  & 11.1 & 10.7 &  9.9  &         9.8            & \textcolor{red}{14.4}   & \textcolor{blue}{12.4}\\
\midrule
AO (\%)          & \textcolor{red}{42.0}   &  31.5 & 31.6 & 32.5 &  29.9 & \textcolor{blue}{34.8} & \textcolor{green}{37.4} &          34.7\\
\bottomrule
\end{tabular}}
\end{center}
\caption{Comparison among nBEKCF-D and state-of-the-art trackers on the GOT10k test set in terms of average overlap (AO) and success rates (SR) at overlap thresholds 0.5 and 0.75. The best three results are shown in red, green and blue, respectively.}
\label{tab:got10k}
\end{table}

\begin{table}
\begin{center}
\resizebox{1\columnwidth}{!}{
\begin{tabular}{|r|c|c|c|c|c|c|c|}
\toprule
Tracker            &        nBEKCF-D          &          UPDT           & ECO  & SiamFC & CFNet & MDNet\\
\midrule
Prec. (\%)         & \textcolor{green}{56.3}  & \textcolor{blue}{55.7}  & 49.2 &  53.3  & 53.3  & \textcolor{red}{56.5}\\
\midrule
Norm.Prec. (\%)    & \textcolor{red}{70.8}    & \textcolor{blue}{70.2}  & 61.8 &  66.6  & 65.4  & \textcolor{green}{70.5}\\
\midrule
Success (AUC) (\%) & \textcolor{red}{61.6}    & \textcolor{green}{61.1} & 55.4 &  57.1  & 57.8  & \textcolor{blue}{60.6}\\
\bottomrule
\end{tabular}}
\end{center}
\caption{Comparison among nBEKCF-D and state-of-the-art trackers on the TrackingNet test set in terms of precision, normalized precision, and success. The best three results are shown in red, green and blue, respectively.}
\label{tab:trackingnet}
\end{table}



\vskip 2mm
{\noindent \bfseries GOT10k}~\cite{huangl18}. We evaluate nBEKCF-D on the test set of GOT10k which is a large-scale tracking benchmark and contains 180 test videos. On the GOT10k experiment, we compare nBEKCF-D against seven state-of-the-art trackers, MDNet~\cite{nam16a}, HCF~\cite{mac15b}, ECO, CCOT, GOTURN~\cite{held16}, SiamFC~\cite{SiamFC-berti16b}, and CFNet~\cite{valmadre17}, in which there is no similar step to SiamRPN to learn to regress the bounding boxes finely. Note that UPDT and fdKCF* are not publicly tested on GOT10k. Following the GOT10k challenge protocol, we evaluate all trackers by average overlap, and success rates at overlap thresholds 0.5 and 0.75. The results are shown in Table~\ref{tab:got10k}. It is seen from the table that nBEKCF-D outperforms other trackers with large margins, except for CFNet in terms of SR$_{0.75}$. In fact, CFNet learns its features in an end-to-end way on large datasets, whereas nBEKCF-D only applies off-the-shelf features.

\vskip 2mm
{\noindent \bfseries TrackingNet}~\cite{muellerm18}. We evaluate our nBEKCF-D on the test set of TrackingNet which is a large-scale tracking benchmark and provides 511 test videos in the wild to assess trackers. On the TrackingNet experiment, we compare nBEKCF-D against five state-of-the-art trackers, MDNet, ECO, SiamFC, CFNet, and UPDT. Identically, there is no similar step to that in SiamRPN to learn to regress the bounding boxes finely in all these trackers. Note that fdKCF* is not publicly tested on TrackingNet. The results are shown in Table~\ref{tab:trackingnet}. It is seen from the table that nBEKCF-D outperforms other trackers, except for MDNet on the precision. It is known that MDNet needs to be trained on large datasets, whereas our nBEKCF need not.

It is interesting to notice from Table~\ref{tab:trackingnet} that our nBEKCF-D still outperforms UPDT, even if it only applies ResNet50 and data augmentation, and does not employ other improvements that UPDT adopted to develop SRDCF.

\vskip 2mm
In summary, it can be seen from the experimental results that avoiding the boundary effect in our nBEKCF improves the accuracy more than just relaxing it in BACF and SRDCF, in the meanwhile, employing non-linear kernels are able to achieve a higher accuracy than employing the linear one.

\section{Conclusions}
\label{sec:conclusion}

The novel kernelized correlation filter without boundary effect is presented in this paper.
Without the boundary effect, being able to exploit non-linear kernels, and running in a high efficiency, nBEKCF possesses all these characteristics that were hard to manage simultaneously before.

Historically, the pioneering work of VanderLugt~\cite{vanderlugt1964} was the start of applying correlation filters to pattern recognition~\cite{kumar2005correlation}. MOSSE and KCF were seminal works in applying correlation filters to visual tracking. Nevertheless, they all were plagued by the well-known defect - the boundary effect. After them, the three representative works, CFLB, SRDCF, and BACF, were proposed to address the defect. While the three alleviated the boundary effect, they lost the merits - high speed and the ability of applying non-linear kernels - of KCF. In this paper, our nBEKCF provides a totally different line to cast off the above dilemma thoroughly, not resorting to FFT. We believe that nBEKCF is an alternative in applying correlation filters to pattern recognition and visual tracking, and the performance of most modern CF trackers would be improved by means of nBEKCF. Even in the era of deep learning, nBEKCF may also provide a theoretical basis for the network design of tracker.



%
%

\section*{Acknowledgment}

This work was supported by National Natural Science Foundation of China under Grants 61976210, 61806200, 61772527 and 61876086, and the Research and Development Projects in the Key Areas of Guangdong Province (Nos.2019B010153001 and 2020B010165001).

\ifCLASSOPTIONcaptionsoff
  \newpage
\fi



%

\bibliographystyle{ieee_fullname}
\bibliography{nBEKCF-latest}

%

%
%
%
%




\end{document}


%
\title{--\textsf{Supplementary Material}--\\ 
\vskip 5mm
Fast Kernelized Correlation Filter\\
without Boundary Effect}
%
%
%

\author{Ming~Tang,~\IEEEmembership{Member,~IEEE,} Linyu~Zheng, Bin Yu,
        and~Jinqiao~Wang,~\IEEEmembership{Member,~IEEE}
\thanks{The authors are with the National Lab of Pattern Recognition, Institute of Automation, Chinese Academy of Sciences (Beijing 100190), and School of Artificial Intelligence, University of Chinese Academy of Sciences (Beijing 100049), China.}
\thanks{The corresponding author: tangm@nlpr.ia.ac.cn.}
}
%
%

\markboth{Journal of \LaTeX\ Class Files,~Vol.~14, No.~8, August~2015}%
{Shell \MakeLowercase{\textit{et al.}}: Bare Demo of IEEEtran.cls for IEEE Journals}
%



\maketitle

\begin{abstract}
This supplementary material contains two parts. 1) An illustration to explain how Algorithm 2, CCIM, works exactly. 2) Mathematical proof of the correctness of CCIM. 
\end{abstract}


%
\IEEEpeerreviewmaketitle

\section{Notation}
\label{sec:notation}
Let $\langle,\rangle$ be the dot product, $\langle\mathbf{H}_1,\mathbf{H}_2\rangle=\langle\text{vec}(\mathbf{H}_1),\text{vec}(\mathbf{H}_2)\rangle$, where $\text{vec}(\mathbf{H})$ indicates the vectorization of the matrix $\mathbf{H}$, $\mathbf{H}_{((a_1,b_1),(a_2,b_2))}$ be the sub-matrix of matrix $\mathbf{H}$ with $(a_1,b_1)$ and $(a_2,b_2)$ as its top left and down right corners, respectively, $H_{a,b}$ or $\mathbf{H}(a,b)$ be an element of matrix $\mathbf{H}$, and $[\bullet]$ is a matrix with $\bullet$ as its element. $\mathbf{F}\text{\footnotesize\FiveStarOpen}\mathbf{S}=
[\langle\mathbf{F},\mathbf{S}_{((a_1,b_1),(a_2,b_2))}\rangle]$,\footnote{In fact, in this paper and its supplementary material, \footnotesize\FiveStarOpen~is exactly the same as the correlation without padding in convolutional neural networks.} where $((a_1,b_1),(a_2,b_2))\in\mathbb{N}^2$ and $\mathbb{N}^2$ is a domain of spatial location. The matrix with a pair of superscripts is still a matrix. 

\section{Illustration of Algorithm 2 (CCIM)}
\label{sec:illustrationsofccim}

\begin{figure*}[ht]
  \centering
  \includegraphics[width=180mm]{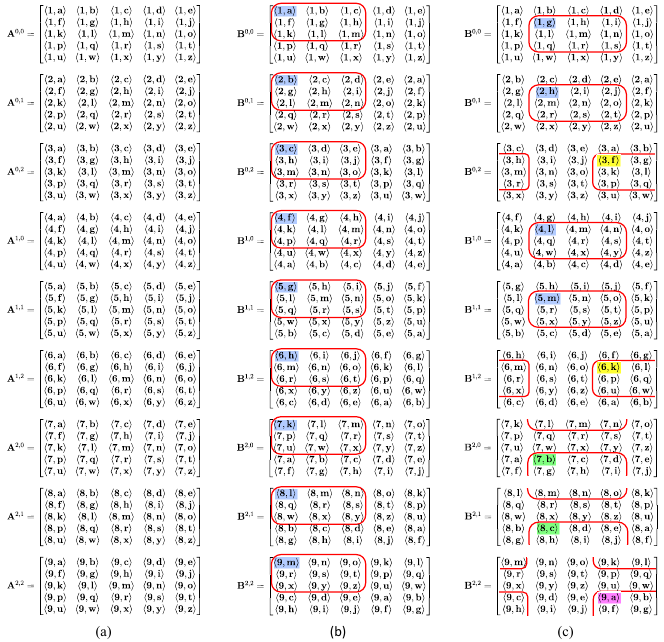}
  \caption{(a) Fundamental calculation matrices $\mathbf{A}^{s,t}$'s. (b) Fundamental matrices, $\mathbf{B}^{s,t}$'s, with sub-matrices marked by red bounding boxes. These sub-matrices are used while calculating $\mathbf{Z}\text{\footnotesize\FiveStarOpen}\mathbf{X}$. See Sec.\ref{sec:corrx00andz} for details. (c) Fundamental matrices, $\mathbf{B}^{s,t}$'s, with sub-matrices marked by red bounding boxes. These sub-matrices are used while calculating $\mathbf{Z}^{1,1}\text{\footnotesize\FiveStarOpen}\mathbf{X}$. See Sec.\ref{sec:corrx11andz} for details.}
\label{fig:matrices}
\end{figure*}

Suppose
\[
\mathbf{Z}\equiv\mathbf{Z^{0,0}}=\left[\begin{array}{ccc}
                         \mathbf{1} & \mathbf{2} & \mathbf{3} \\
                         \mathbf{4} & \mathbf{5} & \mathbf{6} \\
                         \mathbf{7} & \mathbf{8} & \mathbf{9}
                       \end{array}
                 \right]
\]
and
\[
\mathbf{X}=\left[\begin{array}{ccccc}
                   \mathbf{a} & \mathbf{b} & \mathbf{c} & \mathbf{d} & \mathbf{e} \\
                   \mathbf{f} & \mathbf{g} & \mathbf{h} & \mathbf{i} & \mathbf{j} \\
                   \mathbf{k} & \mathbf{l} & \mathbf{m} & \mathbf{n} & \mathbf{o} \\
                   \mathbf{p} & \mathbf{q} & \mathbf{r} & \mathbf{s} & \mathbf{t} \\
                   \mathbf{u} & \mathbf{w} & \mathbf{x} & \mathbf{y} & \mathbf{z}
                 \end{array}
           \right],
\]
where the numbers in $\mathbf{Z}$ indicate its elements, rather than the elements' real values. $\mathbf{z}_{s,t}\in\mathbb{R}^D$ and $\mathbf{x}_{s,t}\in\mathbb{R}^D$ are element of $\mathbf{Z}$ and $\mathbf{X}$, respectively. $\mathbf{Z}$ and $\mathbf{X}$ are called base patch and learning region, respectively.

Let $5\times 5$ fundamental calculation matrices
\[
\mathbf{A}^{s,t}=\mathbf{z}_{s,t}\text{\footnotesize{\FiveStarOpen}}\mathbf{X}
\]
and $5\times 5$ fundamental matrices
\[
\mathbf{B}^{s,t}=\mathbf{P}^{-s}_5\mathbf{A}^{s,t}\mathbf{Q}^{-t}_5=
\mathbf{P}^{-s}_5(\mathbf{z}_{s,t}\text{\footnotesize{\FiveStarOpen}}\mathbf{X})\mathbf{Q}^{-t}_5,
\]
where $\mathbf{P}_5$ and $\mathbf{Q}_5$ are defined in the paper with $m=n=5$, $s=0,1,2$, and $t=0,1,2$. $\mathbf{A}^{s,t}$'s and $\mathbf{B}^{s,t}$'s are shown in Fig.\ref{fig:matrices}.\footnote{All the numbers of equations and figures refer to the equations and figures of this supplementary material.}
Intuitively, $\mathbf{B}^{s,t}$ is generated through cyclically shifting $\mathbf{A}^{s,t}$.

In the rest of this section, we will explain how CCIM works exactly with $\mathbf{Z}\text{\footnotesize\FiveStarOpen}\mathbf{X}$ (\ie, $\mathbf{Z}^{0,0}\text{\footnotesize\FiveStarOpen}\mathbf{X}$) and $\mathbf{Z}^{1,1}\text{\footnotesize\FiveStarOpen}\mathbf{X}$.

\subsection{Correlation of $\mathbf{Z}$ and $\mathbf{X}$}
\label{sec:corrx00andz}
It is seen that $(\mathbf{Z}\text{\footnotesize\FiveStarOpen}\mathbf{X})_{(0,0)}=\langle\mathbf{Z},
\mathbf{X}_{((0,0),(2,2))}\rangle$. According to the construction of $\mathbf{A}^{s,t}$'s and $\mathbf{B}^{s,t}$'s,
\[
\langle\mathbf{z}_{s,t},\mathbf{x}_{s,t}\rangle=B^{s,t}_{0,0},
\]
where $s=0,1,2$ and $t=0,1,2$, $B^{s,t}_{0,0}$ is the element of $\mathbf{B}^{s,t}$ at the top left corner $(0,0)$. $\langle\mathbf{z}_{s,t},\mathbf{x}_{s,t}\rangle$'s are marked by the blue blocks in Fig.\ref{fig:matrices}(b). Therefore,
\[
(\mathbf{Z}\text{\footnotesize\FiveStarOpen}\mathbf{X})_{(0,0)}=\sum_{s=0}^{2}\sum_{t=0}^{2}
\langle\mathbf{z}_{s,t},\mathbf{x}_{s,t}\rangle=\sum_{s=0}^{2}\sum_{t=0}^{2}B^{s,t}_{0,0}.
\]
Generally,
\[
(\mathbf{Z}\text{\footnotesize\FiveStarOpen}\mathbf{X})_{(u,v)}=
\langle\mathbf{Z},
\mathbf{X}_{((u,v),(u+2,v+2))}\rangle,
\]
where
$u=0,1,2$ and $v=0,1,2$. Because
\[
\langle\mathbf{z}_{s,t}, \mathbf{x}_{u+s,v+t}\rangle=B^{s,t}_{u,v},
\]
\[
(\mathbf{Z}\text{\footnotesize\FiveStarOpen}\mathbf{X})_{(u,v)}=\sum_{s=0}^{2}\sum_{t=0}^{2}
\langle\mathbf{z}_{s,t}, \mathbf{x}_{u+s,v+t}\rangle=\sum_{s=0}^{2}\sum_{t=0}^{2}B^{s,t}_{u,v}.
\]
Consequently, 
\[
\mathbf{Z}\text{\footnotesize\FiveStarOpen}\mathbf{X}=
\left(\sum_{s=0}^{2}\sum_{t=0}^{2}\mathbf{B}^{s,t}\right)_{((0,0),(2,2))}.
\]


\subsection{Correlation of $\mathbf{Z}^{1,1}$ and $\mathbf{X}$}
\label{sec:corrx11andz}
\begin{figure}[t]
  \centering
  \includegraphics[height=38mm]{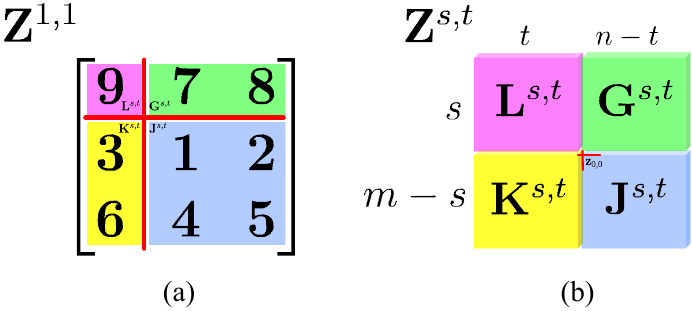}
  \caption{(a) Based on the position of $\mathbf{z}_{0,0}=\mathbf{1}$, $\mathbf{Z}^{1,1}$ is divided into four parts, $\mathbf{L}^{1,1}=(9)$, $\mathbf{G}^{1,1}=(7,8)$, $\mathbf{K}^{1,1}=(3;6)$, and $\mathbf{J}^{1,1}=((1,2);(4,5))$. (b) The division of a general cyclic base $\mathbf{Z}^{s,t}$ into four parts, $\mathbf{L}^{s,t}$, $\mathbf{G}^{s,t}$, $\mathbf{K}^{s,t}$, and $\mathbf{J}^{s,t}$ by means of the position of $\mathbf{z}_{0,0}$.}
\label{fig:divisionofcf}
\end{figure}


$\mathbf{Z}^{1,1}$, as shown in Fig.\ref{fig:divisionofcf}(a), is generated by means of Eq.(\ref{eq:cycliccfs}) defined in Sec.~\ref{sec:correlationofccim} with $m=n=5$ and $s=t=1$. It can be divided into four parts, $\mathbf{L}^{1,1}$, $\mathbf{G}^{1,1}$, $\mathbf{K}^{1,1}$, and $\mathbf{J}^{1,1}$. 
Then, we have
\begin{align*}
\mathbf{Z}^{1,1}\text{\footnotesize\FiveStarOpen}\mathbf{X} & =
(\mathbf{L}^{1,1}\text{\footnotesize\FiveStarOpen}\mathbf{X})_{((0,0),(2,2))}+
(\mathbf{G}^{1,1}\text{\footnotesize\FiveStarOpen}\mathbf{X})_{((0,1),(2,3))}\\
& +
(\mathbf{K}^{1,1}\text{\footnotesize\FiveStarOpen}\mathbf{X})_{((1,0),(3,2))}+
(\mathbf{J}^{1,1}\text{\footnotesize\FiveStarOpen}\mathbf{X})_{((1,1),(3,3))}.
\end{align*}

Suppose the 0$^{\text{th}}$ and 4$^{\text{th}}$ rows are adjacent, and the 0$^{\text{th}}$ and 4$^{\text{th}}$ columns are adjacent too in $\mathbf{B}^{s,t}$'s. Then, $(\mathbf{L}^{1,1}\text{\footnotesize\FiveStarOpen}\mathbf{X})_{((0,0),(2,2))}$ equals to the sub-matrix of $\mathbf{B}^{2,2}$ marked by a red bounding box in $\mathbf{B}^{2,2}$ of Fig.\ref{fig:matrices}(c). Cyclically shifting the marked sub-matrix so as to make the element marked by the pink block in $\mathbf{B}^{2,2}$ of Fig.\ref{fig:matrices}(c) be at the top left corner, and letting $\mathbf{L}$ be the resulting matrix, we have
\[
(\mathbf{L}^{1,1}\text{\footnotesize\FiveStarOpen}\mathbf{X})_{((0,0),(2,2))}=\mathbf{L}_{((0,0),(2,2))}.
\]

Similarly, $(\mathbf{G}^{1,1}\text{\footnotesize\FiveStarOpen}\mathbf{X})_{((0,1),(2,3))}$ equals to the summation of two sub-matrices of $\mathbf{B}^{2,0}$ and $\mathbf{B}^{2,1}$ marked by two red bounding boxes in $\mathbf{B}^{2,0}$ and $\mathbf{B}^{2,1}$ of Fig.\ref{fig:matrices}(c), respectively. Cyclically shifting $\mathbf{B}^{2,0}+\mathbf{B}^{2,1}$ so as to make the summation of two elements marked by the green blocks in $\mathbf{B}^{2,0}$ and $\mathbf{B}^{2,1}$ of Fig.\ref{fig:matrices}(c) be at the top left corner, and letting $\mathbf{G}$ be the resulting matrix, we have
\[
(\mathbf{G}^{1,1}\text{\footnotesize\FiveStarOpen}\mathbf{X})_{((0,1),(2,3))}=\mathbf{G}_{((0,0),(2,2))}.
\]

$(\mathbf{K}^{1,1}\text{\footnotesize\FiveStarOpen}\mathbf{X})_{((1,0),(3,2))}$ equals to the summation of two sub-matrices of $\mathbf{B}^{0,2}$ and $\mathbf{B}^{1,2}$ marked by two red bounding boxes in $\mathbf{B}^{0,2}$ and $\mathbf{B}^{1,2}$ of Fig.\ref{fig:matrices}(c), respectively. Cyclically shifting $\mathbf{B}^{0,2}+\mathbf{B}^{1,2}$ so as to make the summation of two elements marked by the yellow blocks in $\mathbf{B}^{0,2}$ and $\mathbf{B}^{1,2}$ of Fig.\ref{fig:matrices}(c) be at the top left corner, and letting $\mathbf{K}$ be the resulting matrix, we have
\[
(\mathbf{K}^{1,1}\text{\footnotesize\FiveStarOpen}\mathbf{X})_{((1,0),(3,2))}=\mathbf{K}_{((0,0),(2,2))}.
\]

Finally, $(\mathbf{J}^{1,1}\text{\footnotesize\FiveStarOpen}\mathbf{X})_{((1,1),(3,3))}$ equals to the summation of four sub-matrices of $\mathbf{B}^{0,0}$, $\mathbf{B}^{0,1}$, $\mathbf{B}^{1,0}$, and $\mathbf{B}^{1,1}$ marked by four red bounding boxes in $\mathbf{B}^{0,0}$, $\mathbf{B}^{0,1}$, $\mathbf{B}^{1,0}$, and $\mathbf{B}^{1,1}$ of Fig.\ref{fig:matrices}(c), respectively. Cyclically shifting $\mathbf{B}^{0,0}+\mathbf{B}^{0,1}+\mathbf{B}^{1,0}+\mathbf{B}^{1,1}$ so as to make the summation of four elements marked by the blue blocks in $\mathbf{B}^{0,0}$, $\mathbf{B}^{0,1}$, $\mathbf{B}^{1,0}$, and $\mathbf{B}^{1,1}$ of Fig.\ref{fig:matrices}(c) be at the top left corner, and letting $\mathbf{J}$ be the resulting matrix, we have
\[
(\mathbf{J}^{1,1}\text{\footnotesize\FiveStarOpen}\mathbf{X})_{((1,1),(3,3))}=\mathbf{J}_{((0,0),(2,2))}.
\]

Consequently, let $\mathbf{C}^{1,1}=\sum_{\mathbf{D}\in\{\mathbf{L},\mathbf{G},\mathbf{K},\mathbf{J}\}}\mathbf{D}$. Then,
\[
\mathbf{Z}^{1,1}\text{\footnotesize\FiveStarOpen}\mathbf{X}=\mathbf{C}^{1,1}_{((0,0),(2,2))}.
\]

It is easy to see that $\mathbf{Z}^{s,t}\text{\footnotesize\FiveStarOpen}\mathbf{X}$'s can also be obtained with similar procedures, where $s\in\{0,1,2\}$, $t\in\{0,1,2\}$.

\subsection{Design Integral Matrix to Accelerate Correlation}
\label{sec:integralmatrix}
Given $\mathbf{Z}$, according to the construction of $\mathbf{B}^{s,t}$, where $s=0,1,2$ and $t=0,1,2$, $\mathbf{z}_{s,t}$ is the common factor of all elements of the $\mathbf{B}^{s,t}$. Replacing $\mathbf{x}_{s,t}$ by $\mathbf{B}^{s,t}$ in $\mathbf{Z}$, we have the block matrix
\begin{equation*}
  \mathbf{T}=\left[\begin{array}{ccc}
                     \mathbf{B}^{0,0} & \mathbf{B}^{0,1} & \mathbf{B}^{0,2} \\
                     \mathbf{B}^{1,0} & \mathbf{B}^{1,1} & \mathbf{B}^{1,2} \\
                     \mathbf{B}^{2,0} & \mathbf{B}^{2,1} & \mathbf{B}^{2,2}
                   \end{array}
             \right].
\end{equation*}
We call $\mathbf{B}^{s,t}$ a $\mathbf{T}$'s element in the rest of this section.

According to the examples in Secs.\ref{sec:corrx00andz} and~\ref{sec:corrx11andz}, it is necessary to calculate $\sum_{s=0}^{2}\sum_{t=0}^{2}\mathbf{B}^{s,t}$ for obtaining $\mathbf{Z}\text{\footnotesize\FiveStarOpen}\mathbf{X}$, and to calculate $\mathbf{B}^{2,2}$, $\mathbf{B}^{2,0}+\mathbf{B}^{2,1}$, $\mathbf{B}^{0,2}+\mathbf{B}^{1,2}$, and $\mathbf{B}^{0,0}+\mathbf{B}^{0,1}+\mathbf{B}^{1,0}+\mathbf{B}^{1,1}$ for obtaining $\mathbf{Z}^{1,1}\text{\footnotesize\FiveStarOpen}\mathbf{X}$. In order to eliminate repeated additions, the three summations of matrices, which are necessary to calculate both $\mathbf{Z}\text{\footnotesize\FiveStarOpen}\mathbf{X}$ and $\mathbf{Z}^{1,1}\text{\footnotesize\FiveStarOpen}\mathbf{X}$, should be shared. Notice that each above summation of $\mathbf{B}^{s,t}$'s is involved with the summation of $\mathbf{T}$'s elements. Constructing the block integral matrix $\mathbf{M}$, we have
\begin{equation*}
  \mathbf{M}=\left[\begin{array}{ccc}
                     \mathbf{M}^{0,0} & \mathbf{M}^{0,1} & \mathbf{M}^{0,2} \\
                     \mathbf{M}^{1,0} & \mathbf{M}^{1,1} & \mathbf{M}^{1,2} \\
                     \mathbf{M}^{2,0} & \mathbf{M}^{2,1} & \mathbf{M}^{2,2}
                   \end{array}
             \right]
\end{equation*}
with $\mathbf{M}^{i,j}=\sum_{0\leqslant s\leqslant i,0\leqslant t\leqslant j}\mathbf{B}^{s,t}$ as its block elements.\footnote{In the paper, we denote a block element of $\mathbf{M}$ with $\mathbf{M}_{i,j}$.} Then,
\begin{align*}
  &\mathbf{M}^{2,2} = \sum_{s=0}^{2}\sum_{t=0}^{2}\mathbf{B}^{s,t},\\
  &\mathbf{M}^{2,2}-\mathbf{M}^{1,2}-\mathbf{M}^{2,1}+\mathbf{M}^{1,1} = \mathbf{B}^{2,2}, \\
  &\mathbf{M}^{2,1}-\mathbf{M}^{1,1} =\mathbf{B}^{2,0}+\mathbf{B}^{2,1}, \\
  &\mathbf{M}^{1,2}-\mathbf{M}^{1,1} =\mathbf{B}^{0,2}+\mathbf{B}^{1,2},\\
  &\mathbf{M}^{1,1} = \mathbf{B}^{0,0}+\mathbf{B}^{0,1}+\mathbf{B}^{1,0}+\mathbf{B}^{1,1},
\end{align*}
It is seen that the five summations can be achieved within a constant duration by means of $\mathbf{M}$. This way, the repeated additions of $\mathbf{B}^{s,t}$'s are eliminated.


In general, because $\mathbf{D}^{s,t}$ is a sub-matrix of $\mathbf{Z}$, where $\mathbf{D}^{s,t}\in\{\mathbf{L}^{s,t},\mathbf{G}^{s,t},\mathbf{K}^{s,t},\mathbf{J}^{s,t}\}$, the $\mathbf{B}^{i,j}$'s with the elements of $\mathbf{D}^{s,t}$ as their common factors also constitute a $\mathbf{T}$'s elements. While calculating $\mathbf{D}_L^{s,t}$ of Eq.~(\ref{eq:correlationdecomposition}) which is defined in Sec.~\ref{sec:correlationofccim}, similar to the examples in Secs.\ref{sec:corrx00andz} and~\ref{sec:corrx11andz}, it is necessary to calculate the summation of the $\mathbf{T}$'s elements. For a given $\mathbf{Z}$, while $(s,t)$ traverses all possible pairs, \ie, $\mathbf{Z}^{s,t}$ traverses all possible cyclic base patches, the brute-force calculation of the summations will arise a great deal of repeated summations, reducing the efficiency of correlations greatly especially when the size of base patches is large. Therefore, it is a natural choice to construct the integral matrix, as described in CCIM, to obtain such summations without repeated additions.

\section{Correctness of Algorithm 2 (CCIM)}
\label{sec:correlationofccim}
Suppose $\mathbf{Z}\in\mathbb{R}^{m\times n\times D}$ is the base patch with $\mathbf{z}_{s,t}\in\mathbb{R}^D$'s as its elements, cyclic base patches $\mathbf{Z}^{s,t}$'s are generated by using
\begin{equation}
\label{eq:cycliccfs}
\mathbf{Z}^{s,t}(d)=\mathbf{P}_m^{s}\mathbf{Z}(d)\mathbf{Q}_n^{t},
\end{equation}
where $d=0,\ldots,D-1$, $s=0,\ldots,m-1$, and $t=0,\ldots,n-1$,  $\mathbf{Z}(d)$ is the $d$-th channel of $\mathbf{Z}$, $\mathbf{P}_{l_1}$, $l_1\in\{m,M\}$, and $\mathbf{Q}_{l_2}$, $l_2\in\{n,N\}$, are the permutation matrices of $l_1\times l_1$ and $l_2\times l_2$, respectively.\footnote{The detailed explanation of Eq.~(\ref{eq:cycliccfs}) can be found in the paper.}
$\mathbf{X}\in\mathbb{R}^{M\times N\times D}$ is the learning region with $\mathbf{x}_{p,q}\in\mathbb{R}^{D}$'s as its elements.

Let $M\times N$ fundamental calculation matrix
\[
\mathbf{A}^{s,t}=\mathbf{z}_{s,t}\text{\footnotesize{\FiveStarOpen}}\mathbf{X},
\]
where $s=0,\ldots,m-1,t=0,\ldots,n-1$, and $M\times N$ fundamental matrix
\[
\mathbf{B}^{s,t}=\mathbf{P}^{-s}_M\mathbf{A}^{s,t}\mathbf{Q}^{-t}_N=
\mathbf{P}^{-s}_5(\mathbf{z}_{s,t}\text{\footnotesize{\FiveStarOpen}}\mathbf{X})\mathbf{Q}^{-t}_5.
\]
$\mathbf{B}^{s,t}$'s are generated through cyclic-shifts of $\mathbf{A}^{s,t}$'s. Then, $\mathbf{z}_{s,t}$ only appears in $\mathbf{A}^{s,t}$ and $\mathbf{B}^{s,t}$ as a common factor of their elements. The sets of whole $\mathbf{A}^{s,t}$'s and $\mathbf{B}^{s,t}$'s are denoted as $\mathfrak{A}$ and $\mathfrak{B}$, respectively.


In general, as shown in Fig.\ref{fig:divisionofcf}(b), given $s$ and $t$, $\mathbf{Z}^{s,t}$ can always be divided into four parts, $\mathbf{L}^{s,t}$, $\mathbf{G}^{s,t}$, $\mathbf{K}^{s,t}$, and $\mathbf{J}^{s,t}$ by the position of $\mathbf{z}_{0,0}$. The sizes of $\mathbf{L}^{s,t}\equiv[\mathbf{l}^{s,t}_{i,j}]$, $\mathbf{G}^{s,t}\equiv[\mathbf{g}^{s,t}_{i,j}]$, $\mathbf{K}^{s,t}\equiv[\mathbf{k}^{s,t}_{i,j}]$, and $\mathbf{J}^{s,t}\equiv[\mathbf{j}^{s,t}_{i,j}]$ are $s\times t$, $s\times (n-t)$, $(m-s)\times t$, and $(m-s)\times (n-t)$, respectively, and their top left elements are $\mathbf{l}^{s,t}_{0,0}=\mathbf{z}_{m-s,n-t}$, $\mathbf{g}^{s,t}_{0,0}=\mathbf{z}_{m-s,0}$, $\mathbf{k}^{s,t}_{0,0}=\mathbf{z}_{0,n-t}$, and $\mathbf{j}^{s,t}_{0,0}=\mathbf{z}_{0,0}$, respectively. Note that $\mathbf{L}^{s,t}$ and $\mathbf{G}^{s,t}$ or $\mathbf{L}^{s,t}$ and $\mathbf{K}^{s,t}$ will not exist if $s=0$ or $t=0$. $\mathbf{Z}^{s,t}\text{\footnotesize\FiveStarOpen}\mathbf{X}$ can then be divided into four parts, \ie,

\begin{equation}
\label{eq:correlationdecomposition}
  \mathbf{Z}^{s,t}\text{\footnotesize\FiveStarOpen}\mathbf{X}
=\mathbf{D}_{L}^{s,t}+\mathbf{D}_{G}^{s,t}+\mathbf{D}_{K}^{s,t}+\mathbf{D}_{J}^{s,t},
\end{equation}
where
\begin{align*}
\mathbf{D}_{L}^{s,t} & =
  (\mathbf{L}^{s,t}\text{\footnotesize\FiveStarOpen}\mathbf{X})_{((0,0),(M-m,N-n))},\\
\mathbf{D}_{G}^{s,t} & =
  (\mathbf{G}^{s,t}\text{\footnotesize\FiveStarOpen}\mathbf{X})_{((0,t),(M-m,N-n+t))},\\
\mathbf{D}_{K}^{s,t} & =
  (\mathbf{K}^{s,t}\text{\footnotesize\FiveStarOpen}\mathbf{X})_{((s,0),(M-m+s,N-n))},\\
\mathbf{D}_{J}^{s,t} & =
  (\mathbf{J}^{s,t}\text{\footnotesize\FiveStarOpen}\mathbf{X})_{((s,t),(M-m+s,N-n+t))},
\end{align*}
$\mathbf{D}_{L}^{s,t}=[\mathbf{D}_{L}^{s,t}(u,v)]$, $\mathbf{D}_{G}^{s,t}=[\mathbf{D}_{G}^{s,t}(u,v)]$, $\mathbf{D}_{K}^{s,t}=[\mathbf{D}_{K}^{s,t}(u,v)]$, $\mathbf{D}_{J}^{s,t}=[\mathbf{D}_{J}^{s,t}(u,v)]$, $u=0,\ldots,M-m$ and $v=0,\ldots,N-n$.

\subsection{Calculation of $\mathbf{D}_{L}^{s,t}(u,v)$}
\label{sec:dl}

It is clear that
\[
\mathbf{D}_{L}^{s,t}(u,v)=
\sum_{i=0}^{s-1}\sum_{j=0}^{t-1}\langle\mathbf{l}^{s,t}_{i,j},\mathbf{x}_{u+i,v+j}\rangle
\]
and
$\mathbf{l}^{s,t}_{i,j}=\mathbf{z}_{m-s+i,n-t+j}$
which only appears in $\mathbf{B}^{m-s+i,n-t+j}$. We have
\[
\langle\mathbf{l}^{s,t}_{i,j},\mathbf{x}_{u+i,v+j}\rangle =A^{m-s+i,n-t+j}_{u+i,v+j}
\]
and
\begin{align*}
&\mathbf{B}^{m-s+i,n-t+j}=\mathbf{P}_M^{-(m-s+i)}\mathbf{A}^{m-s+i,n-t+j}\mathbf{Q}_N^{-(n-t+j)}\\
&=\mathbf{P}_M^{-(m-s)}\mathbf{P}_M^{-i}\mathbf{A}^{m-s+i,n-t+j}\mathbf{Q}_N^{-j}\mathbf{Q}_N^{-(n-t)}.
\end{align*}
Let
$\mathbf{A}^{(L)}=\mathbf{P}_M^{-i}\mathbf{A}^{m-s+i,n-t+j}\mathbf{Q}_N^{-j}$.
Then,
\[
A^{(L)}_{u,v}=A^{m-s+i,n-t+j}_{u+i,v+j}.
\]
That is,
\[
\langle\mathbf{l}^{s,t}_{i,j},\mathbf{x}_{u+i,v+j}\rangle=A^{(L)}_{u,v}.
\]
\begin{align*}
& \because \mathbf{B}^{m-s+i,n-t+j}=\mathbf{P}_M^{-(m-s)}\mathbf{A}^{(L)}\mathbf{Q}_N^{-(n-t)},\\
& \therefore \langle\mathbf{l}^{s,t}_{i,j},\mathbf{x}_{u+i,v+j}\rangle=
B^{m-s+i,n-t+j}_{(u-(m-s))\text{mod} M,(v-(n-t))\text{mod} N}.
\end{align*}
%
Therefore,
\begin{equation}
\label{eq:D_L}
\mathbf{D}_{L}^{s,t}(u,v)=
\sum_{i=0}^{s-1}\sum_{j=0}^{t-1}B^{m-s+i,n-t+j}_{(u-(m-s))\text{mod} M,(v-(n-t))\text{mod} N}.
\end{equation}
It is seen that the subscripts of every item are irrelative to $(i,j)$ in the right hand side of Eq.~(\ref{eq:D_L}). That is, all of the items possess the same subscripts, although they belong to different fundamental matrices.


It is seen from Fig.\ref{fig:divisionofcf}(b) that the top left and bottom right elements of $\mathbf{L}^{s,t}$ are $\mathbf{z}_{m-s,n-t}$ and $\mathbf{z}_{m-1,n-1}$, respectively. Let $\mathbf{S}^{L,s,t}=\sum_{i=m-s}^{m-1}\sum_{j=n-t}^{n-1}\mathbf{B}^{i,j}$.
According to the construction of integral matrix $\mathbf{M}$, $\mathbf{S}^{L,s,t}$ can be calculated within a constant time as
\begin{align*}
\mathbf{S}^{L,s,t} & = \mathbf{M}^{m-1,n-1}-\mathbf{M}^{m-1,n-t-1} \\
   & -\mathbf{M}^{m-s-1,n-1}+\mathbf{M}^{m-s-1,n-t-1}.
\end{align*}
Then, we have
\[
  \mathbf{D}_{L}^{s,t}(u,v)= S^{L,s,t}_{(u-(m-s))\text{mod} M,(v-(n-t))\text{mod} N}.
\]

\subsection{Calculation of $\mathbf{D}_{G}^{s,t}(u,v)$}
\label{sec:dg}

It is clear that
\[
\mathbf{D}_{G}^{s,t}(u,v)=
\sum_{i=0}^{s-1}\sum_{j=0}^{n-t-1}\langle\mathbf{g}^{s,t}_{i,j},\mathbf{x}_{u+i,v+t+j}\rangle
\]
and
$\mathbf{g}^{s,t}_{i,j}=\mathbf{z}_{m-s+i,j}$
which only appears in $\mathbf{B}^{m-s+i,j}$. We have
\[
\langle\mathbf{g}^{s,t}_{i,j},\mathbf{x}_{u+i,v+t+j}\rangle=A^{m-s+i,j}_{u+i,v+t+j}
\]
and
\begin{align*}
\mathbf{B}^{m-s+i,j} & =\mathbf{P}_M^{-(m-s+i)}\mathbf{A}^{m-s+i,j}\mathbf{Q}_N^{-j}\\
& =\mathbf{P}_M^{-(m-s)}\mathbf{P}_M^{-i}\mathbf{A}^{m-s+i,j}\mathbf{Q}_N^{-j}.
\end{align*}
Let
$\mathbf{A}^{(G)}=\mathbf{P}_M^{-i}\mathbf{A}^{m-s+i,j}\mathbf{Q}_N^{-j}$.
Then,
\[
A^{(G)}_{u,v+t}=A^{m-s+i,j}_{u+i,v+t+j}.
\]
That is,
\[
\langle\mathbf{g}^{s,t}_{i,j},\mathbf{x}_{u+i,v+t+j}\rangle=A^{(G)}_{u,v+t}.
\]
\begin{align*}
& \because \mathbf{B}^{m-s+i,j}=\mathbf{P}_M^{-(m-s)}\mathbf{A}^{(G)},\\
& \therefore \langle\mathbf{g}^{s,t}_{i,j},\mathbf{x}_{u+i,v+t+j}\rangle=
B^{m-s+i,j}_{(u-(m-s))\text{mod} M,v+t}.
\end{align*}
Therefore,
\begin{equation}
\label{eq:D_G}
\mathbf{D}_{G}^{s,t}(u,v)=
\sum_{i=0}^{s-1}\sum_{j=0}^{n-t-1}B^{m-s+i,j}_{(u-(m-s))\text{mod} M,v+t}.
\end{equation}
It is seen that the subscripts of every item are irrelative to $(i,j)$ in the right hand side of Eq.~(\ref{eq:D_G}). That is, all of the items possess the same subscripts, although they belong to different fundamental matrices.

It is seen from Fig.\ref{fig:divisionofcf}(b) that the top left and bottom right elements of $\mathbf{G}^{s,t}$ are $\mathbf{z}_{m-s,0}$ and $\mathbf{z}_{m-1,n-t-1}$, respectively. Let $\mathbf{S}^{G,s,t}=\sum_{i=m-s}^{m-1}\sum_{j=0}^{n-t-1}\mathbf{B}^{i,j}$.
According to the construction of integral matrix, $\mathbf{S}^{G,s,t}$ can be calculated within a constant time as
\[
\mathbf{S}^{G,s,t}=\mathbf{M}^{m-1,n-t-1}-\mathbf{M}^{m-s-1,n-t-1}.
\]
Then, we have
\[
  \mathbf{D}_{G}^{s,t}(u,v)=S^{G,s,t}_{(u-(m-s))\text{mod} M,v+t}.
\]

\subsection{Calculation of $\mathbf{D}_{K}^{s,t}(u,v)$}
\label{sec:dk}
It is clear that
\[
\mathbf{D}_{K}^{s,t}(u,v)=
\sum_{i=0}^{m-s-1}\sum_{j=0}^{t-1}\langle\mathbf{k}^{s,t}_{i,j},\mathbf{x}_{u+s+i,v+j}\rangle
\]
and
$\mathbf{k}^{s,t}_{i,j}=\mathbf{z}_{i,n-t+j}$
which only appears in $\mathbf{B}^{i,n-t+j}$. We have
\[
\langle\mathbf{k}^{s,t}_{i,j},\mathbf{x}_{u+s+i,v+j}\rangle=
A^{i,n-t+j}_{u+s+i,v+j}
\]
and
\begin{align*}
\mathbf{B}^{i,n-t+j} & =\mathbf{P}_M^{-i}\mathbf{A}^{i,n-t+j}\mathbf{Q}_N^{-(n-t+j)}\\
& =\mathbf{P}_M^{-i}\mathbf{A}^{i,n-t+j}\mathbf{Q}_N^{-j}\mathbf{Q}_N^{-(n-t)}.
\end{align*}
Let
$\mathbf{A}^{(K)}=\mathbf{P}_M^{-i}\mathbf{A}^{i,n-t+j}\mathbf{Q}_N^{-j}$.
Then,
\[
A^{(K)}_{u+s,v}=A^{i,n-t+j}_{u+s+i,v+j},
\]
That is,
\[
\langle\mathbf{k}^{s,t}_{i,j},\mathbf{x}_{u+s+i,v+j}\rangle=A^{(K)}_{u+s,v}.
\]
\begin{align*}
& \because \mathbf{B}^{i,n-t+j}=\mathbf{A}^{(K)}\mathbf{Q}_N^{-(n-t)},\\
& \therefore \langle\mathbf{k}^{s,t}_{i,j},\mathbf{x}_{u+s+i,v+j}\rangle=
B^{i,n-t+j}_{u+s,(v-(n-t))\text{mod} N}.
\end{align*}
Therefore,
\begin{equation}
\label{eq:D_K}
\mathbf{D}_{K}^{s,t}(u,v)=
\sum_{i=0}^{m-s-1}\sum_{j=0}^{t-1}B^{i,n-t+j}_{u+s,(v-(n-t))\text{mod} N}.
\end{equation}
It is seen that the subscripts of every item are irrelative to $(i,j)$ in the right hand side of Eq.~(\ref{eq:D_K}). That is, all of the items possess the same subscripts, although they belong to different fundamental matrices.

It is seen from Fig.\ref{fig:divisionofcf}(b) that the top left and bottom right elements of $\mathbf{K}^{s,t}$ are $\mathbf{z}_{0,n-t}$ and $\mathbf{z}_{m-s-1,n-1}$, respectively. Let $\mathbf{S}^{K,s,t}=\sum_{i=0}^{m-s-1}\sum_{j=n-t}^{n-1}\mathbf{B}^{i,j}$.
According to the construction of integral matrix, $\mathbf{S}^{K,s,t}$ can be calculated within a constant time as
\[
\mathbf{S}^{K,s,t}=\mathbf{M}^{m-s-1,n-1}-\mathbf{M}^{m-s-1,n-t-1}.
\]
Then ,we have
\[
  \mathbf{D}_{K}^{s,t}(u,v)=S^{K,s,t}_{u+s,(v-(n-t))\text{mod} N}.
\]

\subsection{Calculation of $\mathbf{D}_{J}^{s,t}(u,v)$}
\label{sec:dj}

It is clear that
\[
\mathbf{D}_{J}^{s,t}(u,v)=
\sum_{i=0}^{m-s-1}\sum_{j=0}^{n-t-1}\langle\mathbf{j}^{s,t}_{i,j},\mathbf{x}_{u+s+i,v+t+j}\rangle
\]
and
$\mathbf{j}^{s,t}_{i,j}=\mathbf{z}_{i,j}$
which only appears in $\mathbf{B}^{i,j}$. We have
\[
\langle\mathbf{j}^{s,t}_{i,j},\mathbf{x}_{u+s+i,v+t+j}\rangle=A^{i,j}_{u+s+i,v+t+j}
\]
and
\[
\mathbf{B}^{i,j}=\mathbf{P}_M^{-i}\mathbf{A}^{i,j}\mathbf{Q}_N^{-j}.
\]
Then, $B^{i,j}_{u+s,v+t}=A^{i,j}_{u+s+i,v+t+j}$.
Therefore
\[
\langle\mathbf{j}^{s,t}_{i,j},\mathbf{x}_{u+s+i,v+t+j}\rangle=B^{i,j}_{u+s,v+t}
\]
and
\begin{equation}
\label{eq:D_J}
\mathbf{D}_{J}^{s,t}(u,v)=
\sum_{i=0}^{m-s-1}\sum_{j=0}^{n-t-1}B^{i,j}_{u+s,v+t}.
\end{equation}
It is seen that the subscripts of every item are irrelative to $(i,j)$ in the right hand side of Eq.~(\ref{eq:D_J}). That is, all of the items possess the same subscripts, although they belong to different fundamental matrices.

It is seen from Fig.\ref{fig:divisionofcf}(b) that the top left and bottom right elements of $\mathbf{J}^{s,t}$ are $\mathbf{z}_{0,0}$ and $\mathbf{z}_{m-s-1,n-t-1}$, respectively. Let $\mathbf{S}^{J,s,t}=\sum_{i=0}^{m-s-1}\sum_{j=0}^{n-t-1}\mathbf{B}^{i,j}$.
According to the construction of integral matrix, $\mathbf{S}^{J,s,t}$ can be calculated within constant time as
\[
\mathbf{S}^{J,s,t}=\mathbf{M}^{m-s-1,n-t-1}.
\]
Then, we have
\[
  \mathbf{D}_{J}^{s,t}(u,v)=S^{J,s,t}_{u+s,v+t}.
\]

\subsection{Correlation of $\mathbf{Z}^{s,t}$ and $\mathbf{X}$}
According to the proofs in Secs.\ref{sec:dl},~\ref{sec:dg},~\ref{sec:dk}, and~\ref{sec:dj}, we have got $\mathbf{D}_{L}^{s,t}$, $\mathbf{D}_{G}^{s,t}$, $\mathbf{D}_{K}^{s,t}$, and $\mathbf{D}_{J}^{s,t}$.
Set
\begin{equation}
\label{eq:increments}
\begin{array}{ll}
r_{L}(u,v) & =(u-(m-s))\text{mod} M - u,\\
c_{L}(u,v) & =(v-(n-t))\text{mod} N - v,\\
r_{G}(u,v) & =(u-(m-s))\text{mod} M - u,\\
c_{G}(u,v) & =t,\\
r_{K}(u,v) & =s,\\
c_{K}(u,v) & =(v-(n-t))\text{mod} N - v,\\
r_{J}(u,v) & =s,\\
c_{J}(u,v) & =t,
\end{array}
\end{equation}
and
\begin{align*}
  \mathbf{S}_L^{s,t} & =\mathbf{P}_M^{-r_{L}(u,v)}\mathbf{S}^{L,s,t}\mathbf{Q}_N^{-c_{L}(u,v)}, \\
  \mathbf{S}_G^{s,t} & =\mathbf{P}_M^{-r_{G}(u,v)}\mathbf{S}^{G,s,t}\mathbf{Q}_N^{-c_{G}(u,v)}, \\
  \mathbf{S}_K^{s,t} & =\mathbf{P}_M^{-r_{K}(u,v)}\mathbf{S}^{K,s,t}\mathbf{Q}_N^{-c_{K}(u,v)}, \\
  \mathbf{S}_J^{s,t} & =\mathbf{P}_M^{-r_{J}(u,v)}\mathbf{S}^{J,s,t}\mathbf{Q}_N^{-c_{J}(u,v)}.
\end{align*}
%
%
Then, $\mathbf{D}_{L}^{s,t}(u,v)$, $\mathbf{D}_{G}^{s,t}(u,v)$,
$\mathbf{D}_{K}^{s,t}(u,v)$, and $\mathbf{D}_{J}^{s,t}(u,v)$ are shifted to $\mathbf{S}_L^{s,t}(u,v)$, $\mathbf{S}_G^{s,t}(u,v)$, $\mathbf{S}_K^{s,t}(u,v)$, and $\mathbf{S}_J^{s,t}(u,v)$, respectively.

Now we prove $\mathbf{P}_M^{-r_{L}(u,v)}=\mathbf{P}_M^{-r_{L}(0,0)}$.

$\because$ $0\leq s\leqslant m-1$, $\therefore$ $-m\leqslant-(m-s)\leqslant-1\leqslant 0$. $\because$ $M\geqslant m$, $\therefore$ $0\leqslant M-m\leq M-(m-s)\leqslant M-1$. $\therefore$ $\mathbf{P}_M^{-r_{L}(0,0)}=\mathbf{P}_M^{-(M-(m-s))}$.

Now we split the set $\{0,\ldots,M-m\}$ into two parts, $S_l=\{0,\ldots,m-s-1\}$ and $S_t=\{m-s,\ldots,M-m\}$.

If $u\in S_l$, then $0\leqslant u\leqslant m-s-1$. $\therefore$ $-M\leqslant-(m-s)\leqslant u-(m-s)\leqslant-1$. In this case, $r_{L}(u,v)=(u-(m-s))\text{mod} M - u=M+u-(m-s)-u=M-(m-s)$.

If $u\in S_t$, then $m-s\leqslant u\leqslant M-m$. $\therefore$ $0\leqslant u-(m-s)\leqslant M-m-(m-s)<M$. In this case, $r^{L}(u,v)=(u-(m-s))\text{mod} M - u=u-(m-s)-u=-(m-s)$.

$\because$ $\mathbf{P}_M^{-(M-(m-s))}=\mathbf{P}_M^{m-s}$, $\therefore$ $\mathbf{P}_M^{-r_{L}(u,v)}=\mathbf{P}_M^{-(M-(m-s))}$, $\therefore$ $\mathbf{P}_M^{-r_{L}(u,v)}=\mathbf{P}_M^{-r_{L}(0,0)}$.

Similarly, we can prove
\begin{align*}
& \mathbf{P}_M^{-r_{G}(u,v)}=\mathbf{P}_M^{-r_{G}(0,0)}, \\
& \mathbf{Q}_N^{-c_{L}(u,v)}=\mathbf{Q}_N^{-c_{L}(0,0)}, \;\;\text{and}\\
& \mathbf{Q}_N^{-c_{K}(u,v)}=\mathbf{Q}_N^{-c_{K}(0,0)}.
\end{align*}

Consequently, given $(s,t)$, the right hands of all equations in Eq.(\ref{eq:increments}) are constant and not related to $(u,v)$. 
All elements of $\mathbf{S}^{L,s,t}$ are shifted by the same number of rows and the same number of columns, respectively. So do $\mathbf{S}^{G,s,t}$, $\mathbf{S}^{K,s,t}$, and $\mathbf{S}^{J,s,t}$. $\mathbf{D}_{L}^{s,t}(0,0)$, $\mathbf{D}_{G}^{s,t}(0,0)$,
$\mathbf{D}_{K}^{s,t}(0,0)$, and $\mathbf{D}_{J}^{s,t}(0,0)$ are shifted to $\mathbf{S}_{L}^{s,t}(0,0)$, $\mathbf{S}_{G}^{s,t}(0,0)$, $\mathbf{S}_{K}^{s,t}(0,0)$, and $\mathbf{S}_{J}^{s,t}(0,0)$, respectively. Specifically,
\begin{align*}
&(\mathbf{S}_L^{s,t})_{((0,0),(M-m-1,N-n-1))}=\\
&\mathbf{S}^{L,s,t}_{((M-m+s,N-n+t),((M-2m+s)\text{mod} M,(N-2n+t)\text{mod} N))},\\
&(\mathbf{S}_G^{s,t})_{((0,0),(M-m-1,N-n-1))}=\\
&\mathbf{S}^{G,s,t}_{((M-m+s,t),((M-2m+s)\text{mod} M,N-n+t))},\\
&(\mathbf{S}_K^{s,t})_{((0,0),(M-m-1,N-n-1))}=\\
&\mathbf{S}^{K,s,t}_{((s,N-n+t),(M-m+s,(N-2n+t)\text{mod} N))},\\
&(\mathbf{S}_J^{s,t})_{((0,0),(M-m-1,N-n-1))}=\\
&\mathbf{S}^{J,s,t}_{((s,t),(M-m+s,N-n+t))}.
\end{align*}
That is, $\mathbf{D}_{L}^{s,t}$, $\mathbf{D}_{G}^{s,t}$, $\mathbf{D}_{K}^{s,t}$, and $\mathbf{D}_{J}^{s,t}$ are aligned.
Let
\[
\mathbf{C}^{s,t}=\mathbf{S}_L^{s,t}+\mathbf{S}_G^{s,t}+
\mathbf{S}_K^{s,t}+\mathbf{S}_J^{s,t}.
\]
According to Eq.(\ref{eq:correlationdecomposition}),
%
\[
\mathbf{Z}^{s,t}\text{\footnotesize\FiveStarOpen}\mathbf{X}=
\mathbf{C}^{s,t}_{((0,0),(M-m-1,N-n-1))}.
\]

\noindent Q.E.D.